\documentclass[default,iicol,sn-nature]{sn-jnl}
\usepackage{graphicx}%
\usepackage{multirow}%
\usepackage{amsmath,amssymb,amsfonts}%
\usepackage{amsthm}%
\usepackage{mathrsfs}%
\usepackage[title]{appendix}%
\usepackage{xcolor}%
\usepackage{textcomp}%
\usepackage{manyfoot}%
\usepackage{booktabs}%
\usepackage{algorithm}%
\usepackage{algorithmicx}%
\usepackage{algpseudocode}%
\usepackage{listings}%
\usepackage{makecell}
\usepackage{multirow}
\usepackage{caption}
\theoremstyle{thmstyleone}%
\theoremstyle{thmstyletwo}%

\theoremstyle{thmstylethree}%

\begin{document}

\title[Article Title]{Towards Ancient Plant Seed Classification: A Benchmark Dataset and Baseline Model}

%%=============================================================%%
%% GivenName	-> \fnm{Joergen W.}
%% Particle	-> \spfx{van der} -> surname prefix
%% FamilyName	-> \sur{Ploeg}
%% Suffix	-> \sfx{IV}
%% \author*[1,2]{\fnm{Joergen W.} \spfx{van der} \sur{Ploeg} 
%%  \sfx{IV}}\email{iauthor@gmail.com}
%%=============================================================%%

\author[1,2]{\fnm{Rui} \sur{Xing}}\email{ruixing@mail.sdu.edu.cn}

\author*[3,4]{\fnm{Runmin} \sur{Cong}}\email{rmcong@sdu.edu.cn}

\author[5,6]{\fnm{Yingying} \sur{Wu}}\email{wyy1577736859@163.com}

\author*[5,6]{\fnm{Can} \sur{Wang}}\email{shandawangcan@sdu.edu.cn}

\author[5,6]{\fnm{Zhongming} \sur{Tang}}\email{tangzm@sdu.edu.cn}

\author[5,6]{\fnm{Fen} \sur{Wang}}\email{wangf@sdu.edu.cn}

\author[5,6]{\fnm{Hao} \sur{Wu}}\email{wharc152@sdu.edu.cn}

\author[7]{\fnm{Sam} \sur{Kwong}}\email{samkwong@ln.edu.hk}
% \equalcont{These authors contributed equally to this work.}

\affil*[1]{\orgdiv{Interdisciplinary Center}, \orgname{Shandong University}, \orgaddress{\city{Jinan}, \postcode{250100}, \state{Shandong}, \country{China}}}

\affil[2]{\orgdiv{Key Scientific Research Base of Paleoenvironment Reconstruction and Subsistence Studies in Archaeology (Shandong University), National Cultural Heritage Administration}, \orgname{Shandong University}, \orgaddress{\city{Qingdao}, \postcode{266237}, \state{Shandong}, \country{China}}}

\affil[3]{\orgdiv{School of Control Science and Engineering}, \orgname{Shandong University}, \orgaddress{\city{Jinan}, \postcode{250061}, \state{Shandong}, \country{China}}}

\affil[4]{\orgdiv{Key Laboratory of Machine Intelligence and System Control}, \orgname{Ministry of Education}, \orgaddress{\city{Jinan}, \postcode{250061}, \state{Shandong}, \country{China}}}

\affil[5]{\orgdiv{Key Laboratory of Archaeological Sciences and Technology (Shandong University)}, \orgname{Ministry of Education}, \orgaddress{\city{Jinan}, \postcode{250100}, \state{Shandong}, \country{China}}}

\affil[6]{\orgdiv{School of Archaeology}, \orgname{Shandong University}, \orgaddress{\city{Jinan}, \postcode{250100}, \state{Shandong}, \country{China}}}

\affil[7]{\orgdiv{School of Data Science}, \orgname{Lingnan University}, \orgaddress{\city{Tuen Mun},  \state{Hong Kong}, \country{China}}}

%%==================================%%
%% Sample for unstructured abstract %%
%%==================================%%

\abstract{
Understanding the dietary preferences of ancient societies and their evolution across periods and regions is crucial for revealing human–environment interactions. Seeds, as important archaeological artifacts, represent a fundamental subject of archaeobotanical research. However, traditional studies rely heavily on expert knowledge, which is often time-consuming and inefficient. Intelligent analysis methods have made progress in various fields of archaeology, but there remains a research gap in data and methods in archaeobotany, especially in the classification task of ancient plant seeds.
To address this, we construct the first Ancient Plant Seed Image Classification (APS) dataset. It contains 8,340 images from 17 genus- or species-level seed categories excavated from 18 archaeological sites across China. In addition, we design a framework specifically for the ancient plant seed classification task (APSNet), which introduces the scale feature (size) of seeds based on learning fine-grained information to guide the network in discovering key “evidence” for sufficient classification. Specifically, we design a Size Perception and Embedding (SPE) module in the encoder part to explicitly extract size information for the purpose of complementing fine-grained information. We propose an Asynchronous Decoupled Decoding (ADD) architecture based on traditional progressive learning to decode features from both channel and spatial perspectives, enabling efficient learning of discriminative features. In both quantitative and qualitative analyses, our approach surpasses existing state-of-the-art image classification methods, achieving an accuracy of 90.5\%. This demonstrates that our work provides an effective tool for large-scale, systematic archaeological research.
}


%%\pacs[JEL Classification]{D8, H51}

%%\pacs[MSC Classification]{35A01, 65L10, 65L12, 65L20, 65L70}

\maketitle

\section{Introduction}%\label{sec1}
\begin{figure*}[ht]
\centering
\includegraphics[width=1\textwidth]{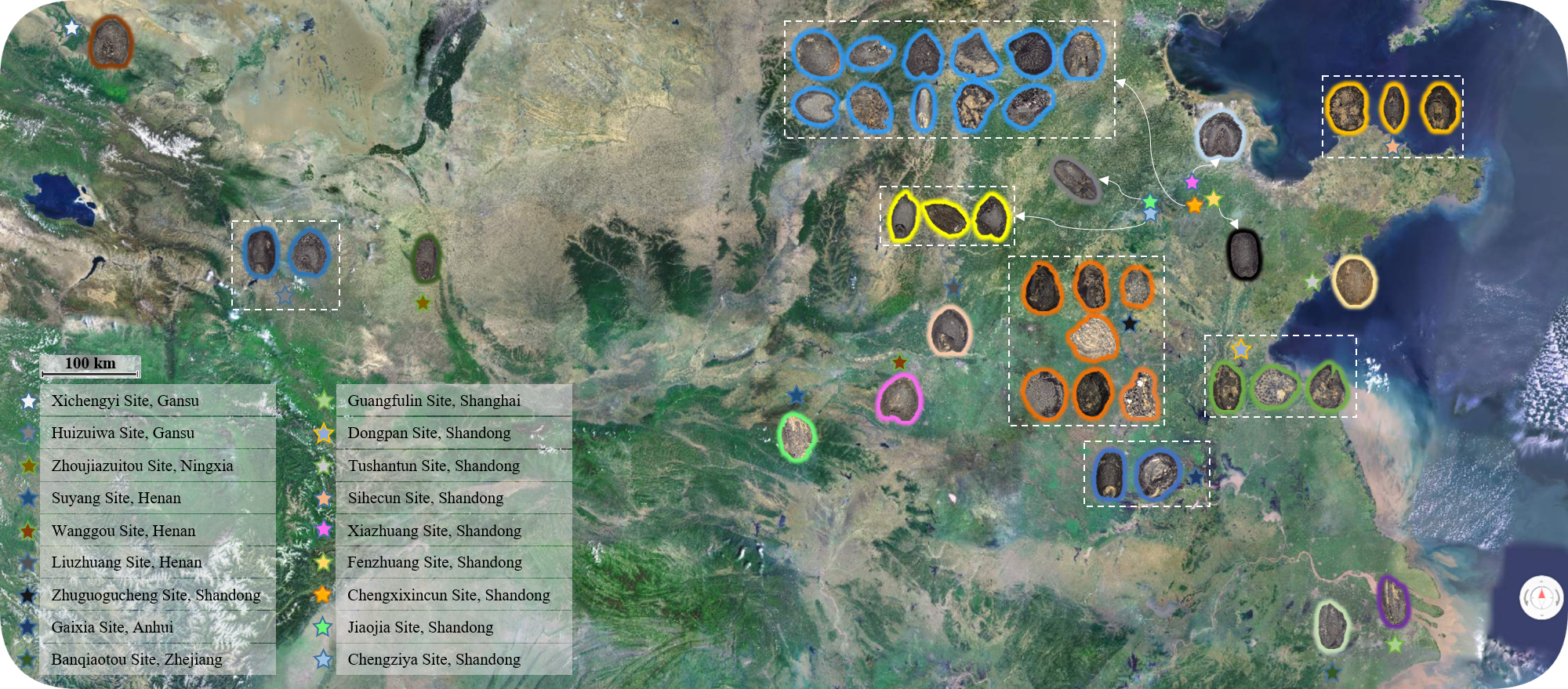}
\caption{Spatial distribution of ancient plant seeds. The figure shows the spatial distribution of the collected categories, covering most regions of ancient China. Stars indicate the locations of archaeological sites, and the seed colors correspond to the temporal distribution shown in Figure \ref{fig2}.}
\label{fig1}
\end{figure*}

China’s agrarian civilization has undergone thousands of years of continuous development, gradually forming a distinctive cultural tradition and social structure \cite{igamberdiev2025human}. Its influence extends beyond state governance and interpersonal ethics, deeply shaping language, theater, folk songs, customs, and ritual practices \cite{tao2023understanding}. Exploring the developmental history of ancient Chinese agriculture holds significant academic value for understanding the origins and evolution of human agricultural systems \cite{wu2025agricultural}. As tangible witnesses of human civilization, archaeological artifacts provide crucial evidence of historical lifeways. Among them, ancient plant seeds, which are direct remnants of ancient agricultural production and dietary patterns, carry essential information about early agricultural practices and have long been central to archaeobotanists research \cite{ramsey2009bayesian}.

However, current research methods remain highly dependent on expert experience, typically involving multiple stages such as excavation, flotation, classification, identification, and analysis. In particular, the classification and analysis of charred seeds require years of specialized training and practical experience for a researcher to become an independent archaeobotany expert \cite{cappers2022manual}. This reliance on manual expertise limits the potential for systematic and quantitative research.

Although in recent years, intelligent archaeological analysis technology has made great progress in the research fields such as mural restoration \cite{shao2023building}, cultural relic dating \cite{zhou2023multi} and fragment splicing \cite{liu2021ams}. However, the intelligent analysis technology in the field of archaeobotany is still in a blank period. The first reason is the lack of large-scale, high-quality datasets to support the training of deep learning models. Therefore, we construct the APS dataset, which can be used for ancient plant seeds classification tasks at the genus- or species-level. To ensure the authenticity and academic value of the data, all samples were collected in a real archaeological research environment and in cooperation with professional archaeobotanists. As shown in Figure \ref{fig1}, these seed specimens date from 5400 BCE to 220 CE and cover 18 archaeological sites in northern and southern China. The dataset includes 17 plant seed categories at the genus- or species-level, including common ancient crop categories (\textit{Hordeum vulgare}, \textit{Triticum aestivum}, \textit{Setaria italica}, \textit{Panicum miliaceum}, etc.) and a category of fruit seeds (\textit{Prunus persica}). To capture clear seed characteristics, we used professional microscopy equipment to collect the side of the seed with the embryo region. In addition, considering the high similarity between categories of ancient seeds, the seed size information is considered as an important factor in the construction of the dataset. We chose to capture seed images in a microscope at 1.6$\times$ magnification to capture their true size in a fixed scene. The proposed dataset poses a greater challenge compared to existing general-purpose image classification datasets: (1) its broad spatio-temporal span and the fragmentation of charred seeds greatly increase the intra-class variation. (2) the high morphological similarity between closely related species, combined with the texture and color changes generated during the charring process, further exacerbates the interclass confusion. (3) the uncertainty of archaeological excavations leads to a severely unbalanced distribution of categories.

% In this study, we aim to establish the first benchmark dataset for archaeobotanical seed classification, enabling systematic evaluation of different models on this task and providing reliable baselines for practical applications. To ensure data authenticity and scientific value, specimens were collected within the actual research environments of archaeologists. As shown in Figure \ref{fig1}, the seed samples span a temporal range from 8000 BCE to 220 CE and cover a wide geographic distribution across northern and southern China. Compared with general-purpose image classification datasets, this dataset presents greater challenges: first, the wide spatiotemporal distribution and the frequent damage to charred seeds lead to substantial intra-class variability; second, close phylogenetic relationships among species, along with the distinctive texture and color characteristics induced by charring, increase inter-class similarity; third, the uncertainty inherent in archaeological excavation results in highly imbalanced class distributions. Together, these factors significantly constrain the performance of conventional image classification methods on this task.

On the other hand, in the subclass partitioning task of morphological approximation, the common paradigm is to learn fine-grained features of the target to achieve classification. For example, recent fine-grained approaches, often based on ViT architectures \cite{dosovitskiy2020image}, attempt to extract discriminative fine-grained features by establishing a pairing relationship between global-local features through fusion \cite{chen2024fet}, or attempt to process features at different granularities of multi-head mechanisms hierarchously to achieve the effect of stepwise search for discriminative features \cite{xu2023fine}. In addition, traditional CNN-based architectures often try to capture more semantic information by expanding the receptive field, and perform detailed learning of richer features through the attention mechanism \cite{he2016deep, han2020ghostnet}. However, the particularity of archaeological scenes (similar charred textures and uncontrollable breakage) prevents fine-grained features from being used as discriminative features alone.  This leads to suboptimal performance of existing methods on ancient plant seed classification tasks. Therefore, we propose an Ancient Plant Seed Classification Network (APSNet), which incorporates size information on the basis of learning fine-grained features to achieve effective classification of genus- or species-level categories. In APSNet, the size information is used in the encoder part to guide the initial learning of the network through the proposed Size Perception and Embedding (SPE) module. Different from the global-local information relationship, we use high-frequency information that is intermediate between edges and masks to characterize the seed size. The advantage of this scale information is that it does not weaken the fine-grained features like Mask, while it does not provide a sparse representation like edge features. Furthermore, traditional methods usually deal with high-level semantic information uniformly during decoding, which cannot adapt to the mixed information extracted at the APSNet encoder stage. Therefore, we design an Asynchronous Decoupled Decoding (ADD) structure in the decoding phase. It achieves the decoupling of mixed information by introducing spatial voting branch on the basis of channel voting branch, and emphasizes the supervision role of spatial information such as size on the network. Extensive experiments show that our model significantly outperforms the current state of the art methods on APS datasets.

Our main contributions are summarized as follows:

\begin{itemize}
\item We compiled 8,340 images of ancient plant seeds spanning 17 categories excavated from 18 archaeological sites across northern and southern China, dating from 5400 BCE to 220 CE. To the best of our knowledge, the APS dataset is the first large-scale image classification dataset for ancient plant seeds, filling a long-standing gap in standardized data resources for archaeobotanical research.

\item We propose APSNet, a classification framework dedicated to ancient plant seeds, which effectively solves the problem of insufficient discriminative feature extraction in traditional methods. The SPE module is proposed to guide the Backbone network to focus on key morphological cues, and the ADD structure is designed to decode high-level semantic information from channel branch and spatial branch simultaneously, which provides additional learning basis for the network in spatial information such as size.

\item We conduct a detailed evaluation of the existing state-of-the-art classification methods on the APS dataset, including 28 classification methods from three different domains (long-tail distribution classification, fine-grained classification, and traditional CNN classification). Experiments demonstrate the effectiveness of the proposed method and establish the first benchmark in the field of ancient seed classification in archaeology, which provides a solid benchmark reference for future research and practical applications.
\end{itemize}

\section{Related Work}%\label{sec2}

% Archaeology employs systematic excavation and multidisciplinary analysis to uncover past human activities and ecosystems, reconstruct the evolution of civilizations, and provide historical insights for contemporary society.  Currently, archaeology remains largely experience-driven, with tasks such as typological comparison and chronological dating relying heavily on individual expert knowledge.  This approach yields low reproducibility and significant discrepancies across research teams.  The process from site excavation to final report often spans several years, making the manual analysis of vast quantities of artifacts a time-consuming and labor-intensive bottleneck.  Plant seed remains, as a prominent example, present significant challenges due to their sheer volume, subtle morphological variations, and scarcity of specialized identifiers.  This severely hampers large-scale, long-term studies of civilizational evolution.

\subsection{Deep Learning in Archaeology}%\label{subsec2.1}

The rapid development of deep learning is providing efficient, accurate and scalable intelligent analysis tools for various branches of archaeology. In artifact identification research, Zhou et al. \cite{zhou2023multi} proposed a learning-based approach that combines deep learning techniques with archaeological knowledge to achieve the task of dating bronze tripods. Resler et al. \cite{resler2021deep} introduced a deep convolutional neural network based on metric learning to identify artifacts by location and period. Jin et al. \cite{jin2025painting} used a convolutional neural network to analyze the distinctive styles of the potters in order to achieve the classification of the painted ceramics.  In fragment splicing research, Liu et al. \cite{liu2021ams} proposed an attention-based multi-scale neural network to extract important geometric and semantic features of terracotta warrior fragments to assist terracotta warrior restoration tasks. Hu et al. \cite{hu2022self} proposed an end-to-end self-supervised model for point cloud data to achieve calibration and segmentation of different parts of intact terracotta warrior data to obtain calibrated fragment data.  In mural restoration research, Shao et al. \cite{shao2023building} proposed an attention diffusion framework and used a specific loss function to deal with the challenging task of mural restoration. Wei et al. \cite{wei2025progressive} proposed a two-stage restoration model to systematically recover identified defects in ancient murals. Zeng et al. \cite{zeng2024virtual} used a convolutional neural network and a matching image block-based approach, respectively, to restore images based on the size of the area to be restored in historical murals.  In the study of ancient writing and inscription interpretation, Yuan et al. \cite{yuan2022extracting} combined deep learning and natural language processing techniques to investigate a text information extraction method for automatically obtaining spatio-temporal information of sites. Zhou et al. \cite{zhou2025oraclenet} introduced an image processing model for oracle bone characters to realize oracle bone character recognition.

Although deep learning has made progress in archaeological subfields such as bronze dating, fragment splicing, mural restoration and ancient text recognition, there is still a lack of publicly available datasets and effective methods in the face of the huge number and highly similar morphology of plant seed macro-remains.

\subsection{Image Classification}%\label{subsec2.2}

\begin{figure*}[h]
\centering
\includegraphics[width=1\textwidth]{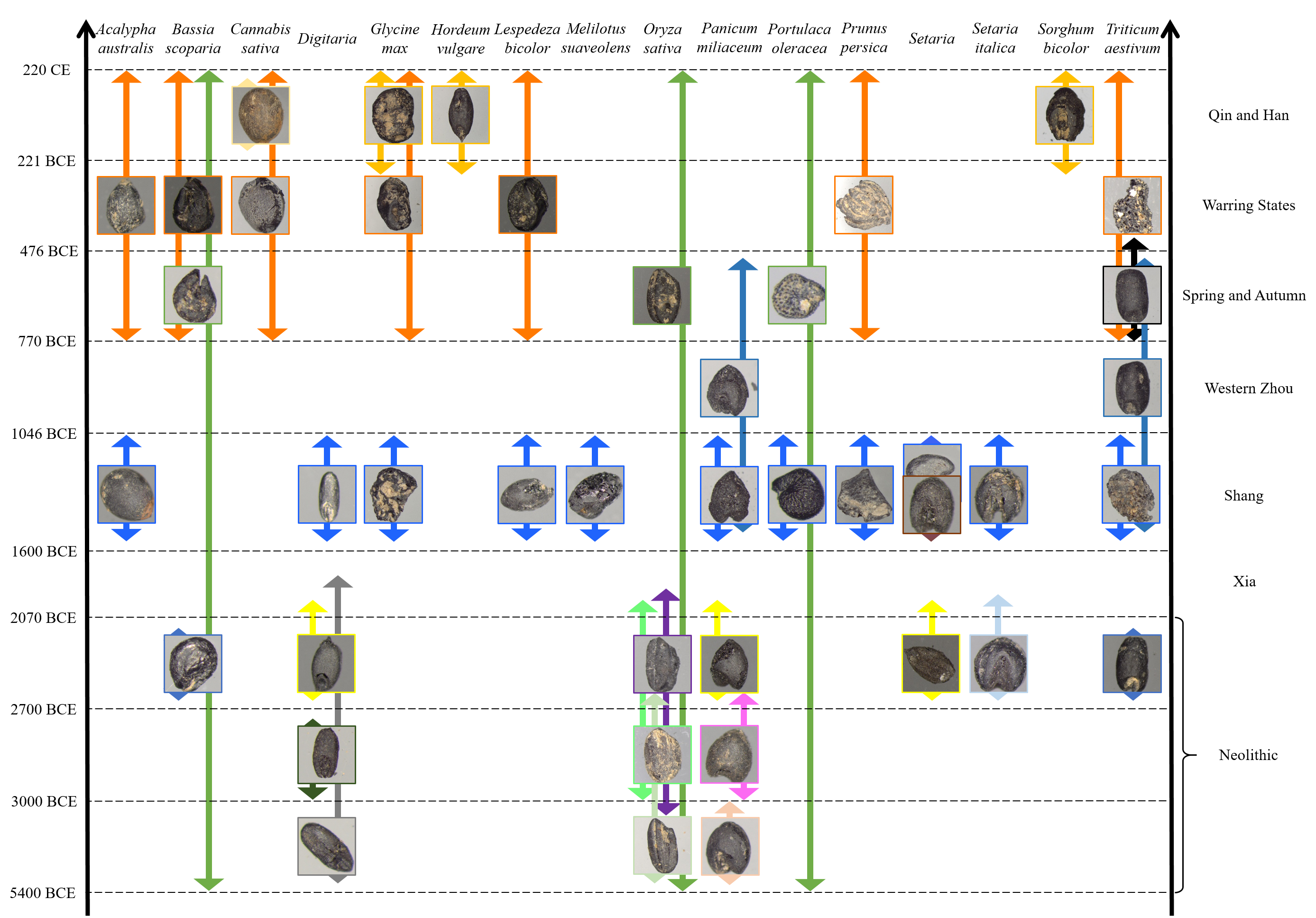}
\caption{Temporal distribution of ancient plant seeds.   The figure shows the temporal distribution of the ruins to which the different seeds belong, spanning from 5400 BCE to 220 CE.   BCE denotes “Before Common Era” and CE denotes “Common Era”.}
\label{fig2}
\end{figure*}

Since the introduction of the pioneering deep neural network LeNet \cite{lecun2002gradient}, universal visual Backbone architectures for image classification tasks have undergone continuous evolution. The emergence of networks such as AlexNet \cite{krizhevsky2012imagenet}, VGGNet \cite{simonyan2014very}, GoogLeNet \cite{szegedy2015going}, and ResNet \cite{he2016deep} has progressively driven improvements in image classification performance and effectiveness. ConvNeXt \cite{liu2022convnet}, EfficientNet \cite{tan2019efficientnet, tan2021efficientnetv2}, Swin \cite{liu2021swin}, and ViT \cite{dosovitskiy2020image} pushed ImageNet \cite{deng2009imagenet} Top-1 accuracy beyond 88\% through global-local coupling. MobileNet \cite{howard2017mobilenets, sandler2018mobilenetv2, howard2019searching}, ShuffleNet \cite{zhang2018shufflenet, ma2018shufflenet}, and the GhostNet \cite{han2020ghostnet, tang2022ghostnetv2, liu2024ghostnetv3} series compressed parameters to under 5 million through batch convolutions and linear bottlenecks, achieving 0.3–0.5 GFLOPS on 224$\times$224 inputs to enable millisecond-level edge inference. For long-tail scenarios, LOS \cite{sunrethinking} and LTR \cite{alshammari2022long} employ dynamic reweighting and meta-resampling, boosting tail-class recall by 8–12\% and effectively mitigating extreme imbalance. For fine-grained tasks, FET-FGVC \cite{chen2024fet}, AA-Trans \cite{wang2023aa}, HAVT \cite{hu2023hierarchical}, and IELT \cite{xu2023fine} introduce component-global dual branches, attention token refinement, and hierarchical visual transformers, boosting baselines by 4–7\% on datasets like CUB-200-2011 \cite{WahCUB_200_2011} and FGVC-Aircraft \cite{maji2013fine}. SeaFormer \cite{wan2025seaformer++} and TCFormer \cite{zeng2024tcformer} further maintain linear complexity through a coexistent convolution-attention structure, while 2DMamba \cite{zhang20252dmamba} achieves O(n) global modeling on 256$\times$256 inputs via a state space mechanism, establishing a new paradigm for classification demands with weak textures, few samples, and high similarity. Although these models demonstrate strong performance in general, long-tail, and fine-grained domains respectively, their robustness against the triple challenges of “high intra-class variability, low inter-class variability, and extremely skewed distributions” in archaeological plant remains has yet to be validated.

In summary, while deep learning has achieved remarkable success in areas such as artifact restoration, object recognition, and chronological estimation, plant analysis remains a weak link in the field. There is an urgent need for large-scale, high-quality datasets of ancient plant seeds and specialized analytical methods to provide an intelligent and quantitative foundation for broader archaeological research topics, such as the origins of agriculture and ecological adaptation.

\section{APS Dataset}%\label{sec3}

In this study, we introduce the APS dataset, which includes most common genus or species level categories and is specifically designed for the classification of ancient plant seeds. In current archaeobotanical research, the identification of ancient seeds largely relies on expert knowledge, and the lack of dedicated data resources limits the application of deep learning methods in this field. Therefore, there is an urgent need for a standardized dataset that reflects real archaeological research contexts and encompasses plant seeds from major historical sites. This section describes the construction of the APS dataset, including the selection of data collection regions, the imaging equipment used, and the criteria for defining sample categories.

\begin{figure*}[h]
\centering
\includegraphics[width=1\textwidth]{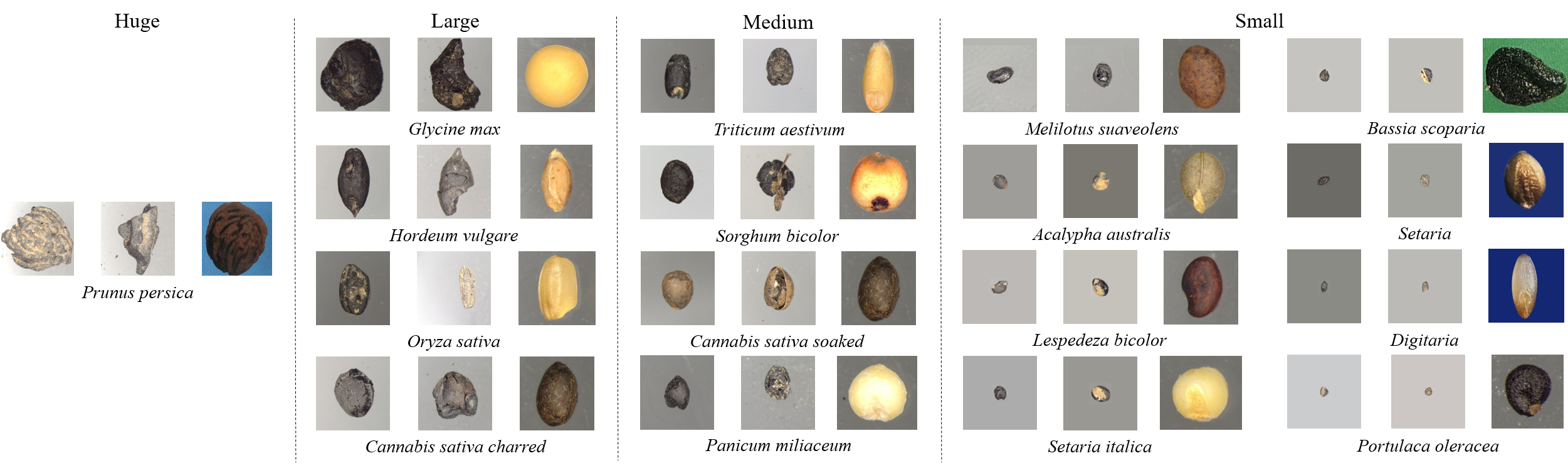}
\caption{Comparison of the size and damage of ancient plant seeds and the condition of modern seeds. We divide the seeds into four sizes, where each has three columns of seed images.  The first column shows the true seed size under 1.6$\times$ magnification, the second column shows seeds with pronounced differences caused by uncontrollable factors, and the third column shows the corresponding modern seeds.}
\label{fig3}
\end{figure*}

\subsection{Preparation}%\label{subsec3.1}

\subsubsection{Selection of the Data Collection Domain}%\label{subsubsec3.1.1}

Current archaeobotanical research often focuses on crop remains from a single region or a limited time period to investigate historical developments within specific eras. While this approach has yielded important insights into local agricultural evolution, it is limited in supporting systematic and efficient classification of plant seeds across the full temporal span of history. Computer vision offers a promising solution for such tasks, but datasets covering the broad spatiotemporal distribution of ancient plant seeds remain scarce. As shown in Figures \ref{fig1} and \ref{fig2}, we collected plant remains from 18 archaeological sites distributed across northern and southern China, spanning a temporal range from 5400 BCE to 220 CE.

\subsubsection{High Resolution Microscope Camera Equipment}%\label{subsubsec3.1.2}

To obtain sample images consistent with actual archaeological research conditions, we captured specimens using an OLYMPUS SC180 digital microscope camera after completing pre-processing procedures such as cleaning and flotation. The device features advanced color calibration technology, enabling high-fidelity and accurate color reproduction while ensuring consistency across input and output as well as between different cameras. In our experiments, a 1/2.3-inch CMOS color image sensor was used, with a resolution of 4912 $\times$ 3684 pixels and a pixel size of 1.25 $\times$ 1.25 $\mu$m. Images were transmitted to a computer via USB 3.0 at 10.5 FPS for visualization. Overall, the OLYMPUS SC180 integrates a high-resolution image sensor, microscope imaging optimization algorithms, and professional imaging interfaces, providing high-quality imaging and precise documentation for life sciences, pathology, materials research, and the archaeological applications in this study.

\subsubsection{Collection Target and Class Definition}%\label{subsubsec3.1.3}

Genus- or species-level classification requires integrating both morphological and textural information of the target.  Therefore, during image acquisition, we aimed to capture the seed front that conveys the most information.  As shown in Figure \ref{fig3}, the charring process often diminishes surface texture, and long-term preservation can cause damage, leading to inconsistent shapes.  These factors blur inter-class differences, making reliance solely on shape and texture insufficient for neural networks to learn stable discriminative features.

Based on the experience of archaeobotanical experts, we introduced size under a 1.6$\times$ magnification as an additional feature to enhance inter-species distinguishability. The \textit{Prunus persica} represents the seed of a fruit and is morphologically much larger than other categories; thus, we defined it as filling the entire image, similar to \textit{Glycine max} in relative size.  \textit{Cannabis sativa} has two preservation conditions (soaked and charred), and the resulting morphological differences were sufficiently large to define them as two separate categories.

It is important to note that the introduction of size only partially alleviates ambiguities in texture and shape; for most categories, texture remains a crucial discriminative feature, posing a greater challenge for model classification. Additionally, Appendix \ref{secA1} presents the classification criteria of the APS dataset, with genus or species level definitions in archaeobotany serving as our primary reference.

\subsection{Dataset Construction and Statistics}%\label{subsec3.2}

\begin{table}[t]
\caption{\centering Raw Dataset Information}\label{tab1}
\centering
\begin{tabular*}{\linewidth}{@{}lccc@{}}
\toprule
Quantity & \makecell[c]{Number of\\ Classes}  & Resolution & \makecell[c]{Whether to\\ Introduce Size\\ Differences}\\
\midrule
8986    & 23   & 4912 $\times$ 3684  & No  \\
\botrule
\end{tabular*}
% \footnotetext{Source: This is an example of table footnote. This is an example of table footnote.}
% \footnotetext[1]{Example for a first table footnote. This is an example of table footnote.}
% \footnotetext[2]{Example for a second table footnote. This is an example of table footnote.}
\end{table}

\subsubsection{Data acquisition and preprocessing}%\label{subsubsec3.2.1}

\begin{table*}[h]
\caption{\centering APS Dataset Structure}\label{tab2}
\centering
\resizebox{0.8\textwidth}{!}{
\begin{tabular}{@{}lccccc@{}}
\toprule
Classes & \makecell[c]{Classification\\ level} & \makecell[c]{Train set\\ number} & \makecell[c]{Test set\\ number} & Resolution & \makecell[c]{Whether to\\ Introduce Size\\ Differences}\\
\midrule
\textit{Acalypha australis} & Species & 25 & 20 & 512 $\times$ 512 & Yes \\
\textit{Bassia scoparia} & Species & 25 & 21 & 512 $\times$ 512 & Yes \\
\textit{Cannabis sativa charred} & Species & 145 & 137 & 512 $\times$ 512 & Yes \\
\textit{Cannabis sativa soaked} & Species & 515 & 193 & 512 $\times$ 512 & Yes \\
\textit{Digitaria} & Genus & 637 & 125 & 512 $\times$ 512 & Yes \\
\textit{Glycine max} & Species & 384 & 335 & 512 $\times$ 512 & Yes \\
\textit{Hordeum vulgare} & Species & 332 & 141 & 512 $\times$ 512 & Yes \\
\textit{Lespedeza bicolor} & Species & 72 & 13 & 512 $\times$ 512 & Yes \\
\textit{Melilotus suaveolens} & Species & 40 & 14 & 512 $\times$ 512 & Yes \\
\textit{Oryza sativa} & Species & 491 & 245 & 512 $\times$ 512 & Yes \\
\textit{Panicum miliaceum} & Species & 580 & 345 & 512 $\times$ 512 & Yes \\
\textit{Portulaca oleracea} & Species & 50 & 7 & 512 $\times$ 512 & Yes \\
\textit{Prunus persica} & Species & 42 & 3 & 512 $\times$ 512 & No \\
\textit{Setaria} & Genus & 220 & 76 & 512 $\times$ 512 & Yes \\
\textit{Setaria italica} & Species & 1245 & 479 & 512 $\times$ 512 & Yes \\
\textit{Sorghum bicolor} & Species & 171 & 103 & 512 $\times$ 512 & Yes \\
\textit{Triticum aestivum} & Species & 866 & 243 & 512 $\times$ 512 & Yes \\
\botrule
\end{tabular}}
% \footnotetext{Source: This is an example of table footnote. This is an example of table footnote.}
% \footnotetext[1]{Example for a first table footnote. This is an example of table footnote.}
% \footnotetext[2]{Example for a second table footnote. This is an example of table footnote.}
\end{table*}

This section describes the acquisition, processing, and construction of the original images in the APS dataset. Table \ref{tab1} presents the basic information of the original dataset. The raw images were collected collaboratively by 16 archaeobotanical experts and computer vision specialists. The dataset comprises 23 genus- or species-level categories of ancient plant seeds, totaling 8,986 images with a resolution of 4912 $\times$ 3684 pixels. The class with the largest number of images is \textit{Setaria italica}, containing 1,778 images, whereas the class with the fewest images is \textit{Zizyphus jujube}, with 12 images. To accommodate the requirements of mainstream neural network training and improve processing efficiency, all images were downsampled to 512 $\times$ 512 using bilinear interpolation. 
Subsequently, the raw data were manually filtered according to the class definitions in Appendix \ref{secA1}, excluding seven categories deemed too severely damaged to provide training value.

\begin{figure}[h]
\centering
\includegraphics[width=1\linewidth]{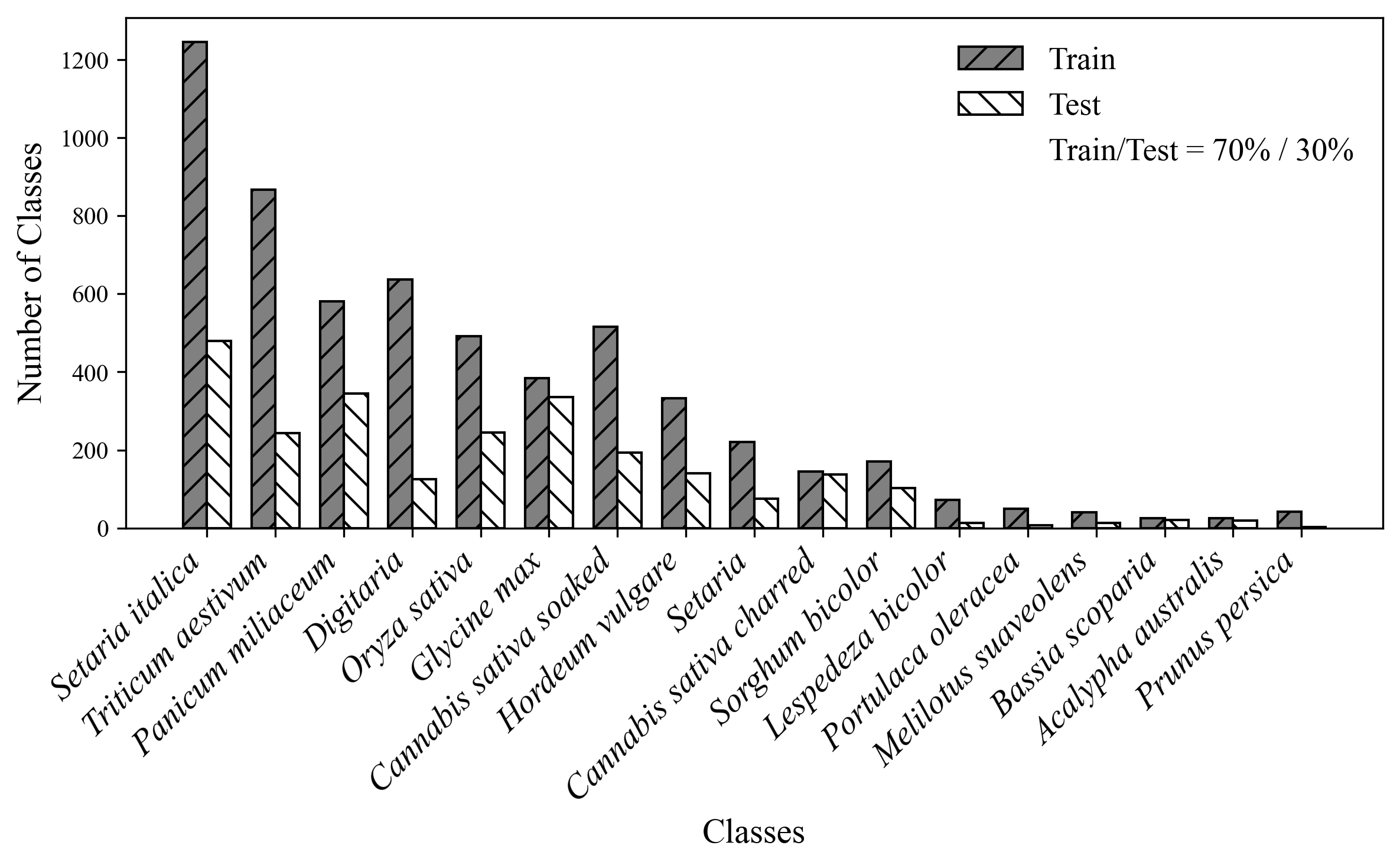}
\caption{Visualization of the statistical distribution of the APS dataset. The x-axis represents the class and the y-axis represents the number of samples.}
\label{fig4.1}
\end{figure}

\subsubsection{Data Statistics and Analysis}%\label{subsubsec3.2.2}

\begin{figure*}[h]
\centering
\includegraphics[width=1\textwidth]{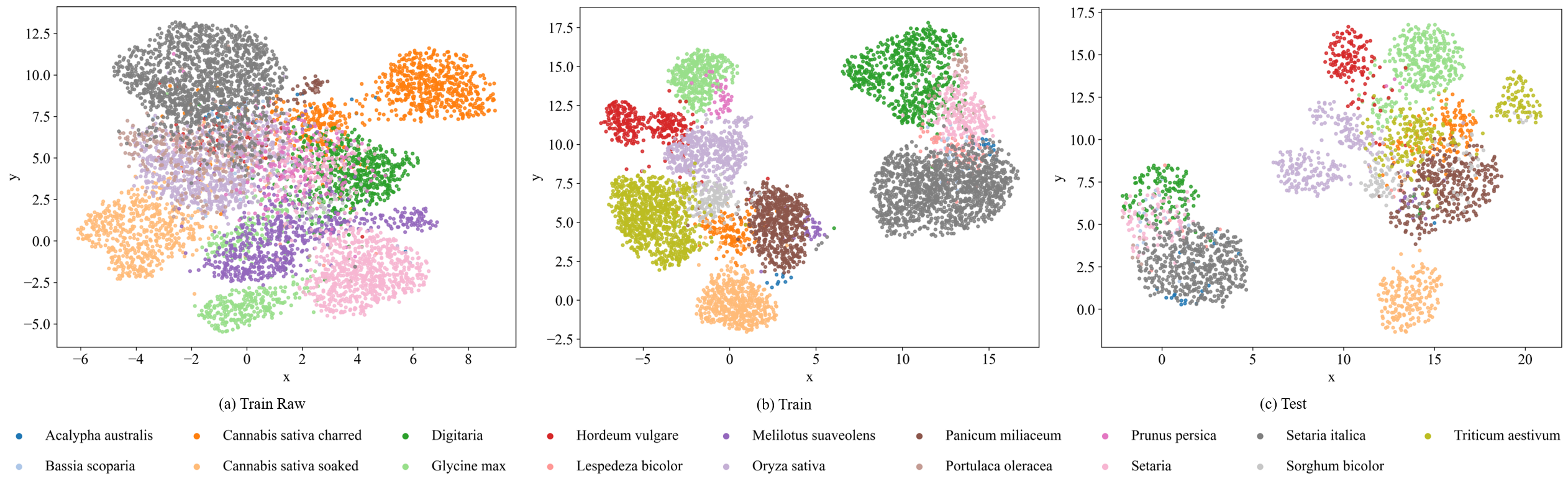}
\caption{Visualization of feature distribution of the dataset. The x- and y-axes represent spatial positions.}
\label{fig4.2}
\end{figure*}

We also provide a detailed analysis and discussion of the APS dataset. Table \ref{tab2} presents the structure of the APS dataset, which comprises 17 categories—including one fruit seed, the \textit{Prunus persica}—and a total of 8,340 images. Each image contains a single seed centrally positioned. The dataset was split into training and test sets at a 7:3 ratio. Figure \ref{fig4.1} shows the distribution of images across categories, revealing a clear long-tail pattern in the APS dataset. To better reflect the randomness of sample distribution in real experiments, we did not enforce a fixed proportion of categories between the training and test sets.

Figure \ref{fig4.2}(a) illustrates the feature distribution of the training set when size information is not included, using features extracted from a ResNet50 network pretrained on ImageNet1K. It can be observed that, without size information, the training set exhibits small inter-class differences and large intra-class variations. We argue that this characteristic limits the APS dataset’s ability to provide effective discriminative features for neural network training and should be addressed during data preprocessing. The practical performance further supports this point: APS datasets without size information achieve high accuracy only within closed sets and show poor generalization. Moreover, the irregular feature distribution highlights the significant differences between the APS dataset and large-scale general datasets such as ImageNet, emphasizing the necessity of the APS dataset proposed in this study.

Figures \ref{fig4.2}(b) and \ref{fig4.2}(c) display the feature distributions of the training and testing sets after incorporating size information in the ResNet50 network. The final APS dataset exhibits a reasonable feature distribution that can effectively support the training process. Notably, although the introduction of size information improves performance, most categories still require classification methods to capture fine-grained details of texture, shape, and color. Therefore, genus or species level recognition of ancient charred plant seeds remains a highly challenging task.

\section{Ancient Plant Seed Image Classification Network}%\label{sec4}

\begin{figure*}[h]
\centering
\includegraphics[width=1\textwidth]{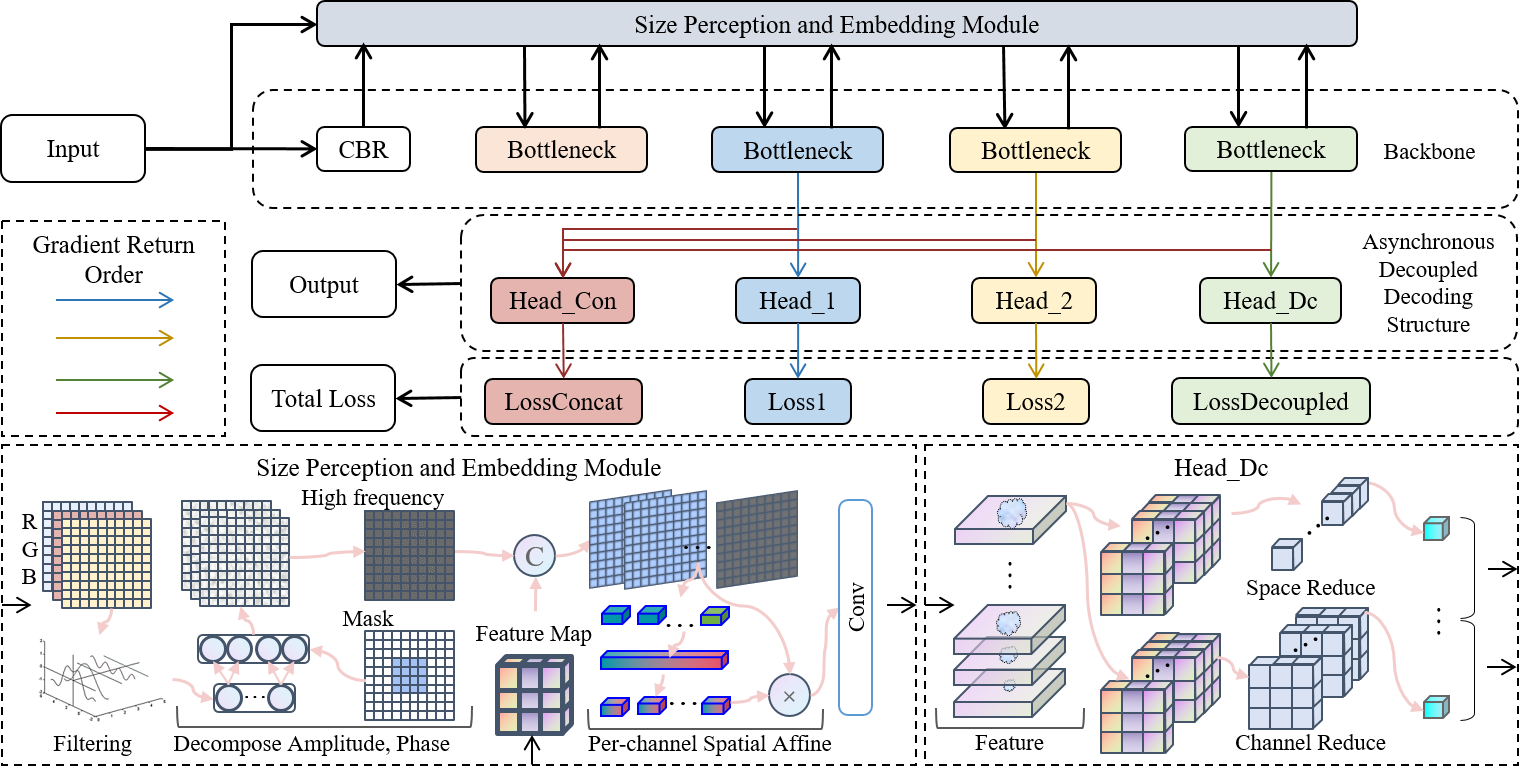}
\caption{APSNet Architecture.   The Head\_Con classification head is obtained by concatenating Head\_1, Head\_2 and Head\_Dc.   The CBR module denotes the cascade of Conv, BatchNorm, and ReLU.   The LossConcat, Loss1 and Loss2 is the cross-entropy loss function, LossComposite is the sum of cross-entropy loss function and contrastive loss.}\label{fig5}
\end{figure*}

% Aiming at the problems of high inter-class similarity, large intra-class differences and serious class imbalance in the classification task of ancient plant seeds, we propose a deep learning architecture—APSNet, which is capable of automatically perceiving size and texture details features. As illustrated in Figure \ref{fig5}, APSNet consists of three main components: the Size Perception Module (SPE), the Backbone network, and the Asynchronous Decoupled Decoding (ADD) structure. The core of the SPE module lies in the Size Perception Modeling (SPM), which applies Fourier transform to capture high-frequency information from the images. We use this information containing outline to explicitly characterize ancient seed size, while the emphasis on detailed texture information helps to establish the relationship between size and fine-grained features.  Feature Channel Affine fuses this high-frequency information into the backbone through channel-by-channel affine, guiding the backbone to automatically perceive the size information based on learning the detailed texture.

% After the backbone, the ADD Classification structure learns the Multi-Granularity features through four different stages of training. Among them, we use the dual-domain voting mechanism when Head\_Dc classifies the head, and introduce the spatial domain voting from the channel domain voting to explicitly conduct comparative supervision of the size spatial information.

Whether it is the fusion of global-local features \cite{chen2024fet}, the hierarchical processing of the decoder \cite{xu2023fine}, or the acquisition of rich information by a large receptive field \cite{he2016deep, han2020ghostnet}, all of these solutions uniformly treat the discriminative features of seeds, resulting in inefficient classification performance of ancient plant seeds. Although global-local feature fusion attempts to obtain useful information from different perspectives, it is not helpful for ancient plant seed objects. This is because the singleness of the observation scene leads to global information approximation, and the size of drastic changes leads to excessive local information difference. Therefore, this fusion method essentially still does not find discriminative information representation. According to the experience of archaeologists, a tried-and-true scheme is to first distinguish approximate categories by size, and then determine the final class based on fine-grained features such as embryonic regions or corners.

Therefore, we propose APSNet, a deep learning architecture that can automatically perceive size and texture detail features. As shown in Figure \ref{fig5}, APSNet first adopts the newly designed Size Perception and Embedding (SPE) module in the encoding part. The SPE module introduced the size-aware ability on the basis of the original learning ability by extracting a kind of seed size information and embedding it into the Backbone network. In order to further distinguish categories by fine-grained features after learning the size, we propose an asynchronous decoupled decoding (ADD) structure in the decoding stage. We perform decoding from both channel and spatial perspectives to achieve subtle discrimination of size approximate seeds.

\subsection{SPE Module}\label{subsec4.1}

The role of the SPE module is to guide the backbone network to learn size features by extracting scale-related information. Different from the implicit calculation method of spatial attention whose complexity is $O(n^{2})$, we choose the linear complexity of Fourier transform to extract high-frequency components as explicit prior information. At the same time, different from the global-local feature fusion method, the SPE module mainly focuses on capturing the information in the local features that can represent the dimension. This design choice is based on the observation that slowly varying global features exhibit high similarity at the micro level, making them less effective as guiding information during training. In addition, how to effectively embed the prior information into the Backbone network becomes a key step in the encoding stage. We argue that simple operations, such as addition, are easily affected by irrelevant regions. Therefore, we integrate the size information by a channel-by-channel fusion mechanism.

Firstly, extracting scale information that can effectively represent the size is the core of SPE. Generally, the spatial attention mechanism \cite{Woo_2018_ECCV} is used to compute useful information about spatial relationships. However, this way is an implicit expression in extracting scale information: noise and other interference information may be captured in shallow layers, and detailed textures are often preferred in deep layers. Therefore, we choose to extract a kind of prior information from the shallow layer and pass it layer by layer. This practice of explicitly guiding the Backbone network to sense size helps to maintain a consistent representation in shallow and deep layers.

We analyze different prior information of ancient plant seed images from two perspectives: information \cite{6773024, 477498} and feature. In the case of \textit{Acalypha australis}, we randomly selected 10 images to evaluate the mean Shannon entropy of different prior information. Among them, the original image is about 0.9 bit, the High-frequency is about 1.4 bit, the Mask is about 0.3 bit, and the Edge is about 0.1 bit. The reason that the information entropy of high-frequency is the highest is that it magnifies the drastic changes of seed features relative to the original image. Compared to Mask and Edge, it preserves the edges and details of the seed. Does this mean that rich high-frequency information is more appropriate to describe the size of seeds? To this end, we further analyze in terms of features. We compare the feature quality of different priors at different layers of the network to examine how they perform at deeper layers (see Section \ref{subsubsec5.3.2} and \ref{subsubsec5.4.2}). In summary, after analysis, we believe that high-frequency information can maintain the effect in deeper layers of the network and effectively guide Backbone to perceive the seed size.

\begin{figure}[t]
\centering
\includegraphics[width=1\linewidth]{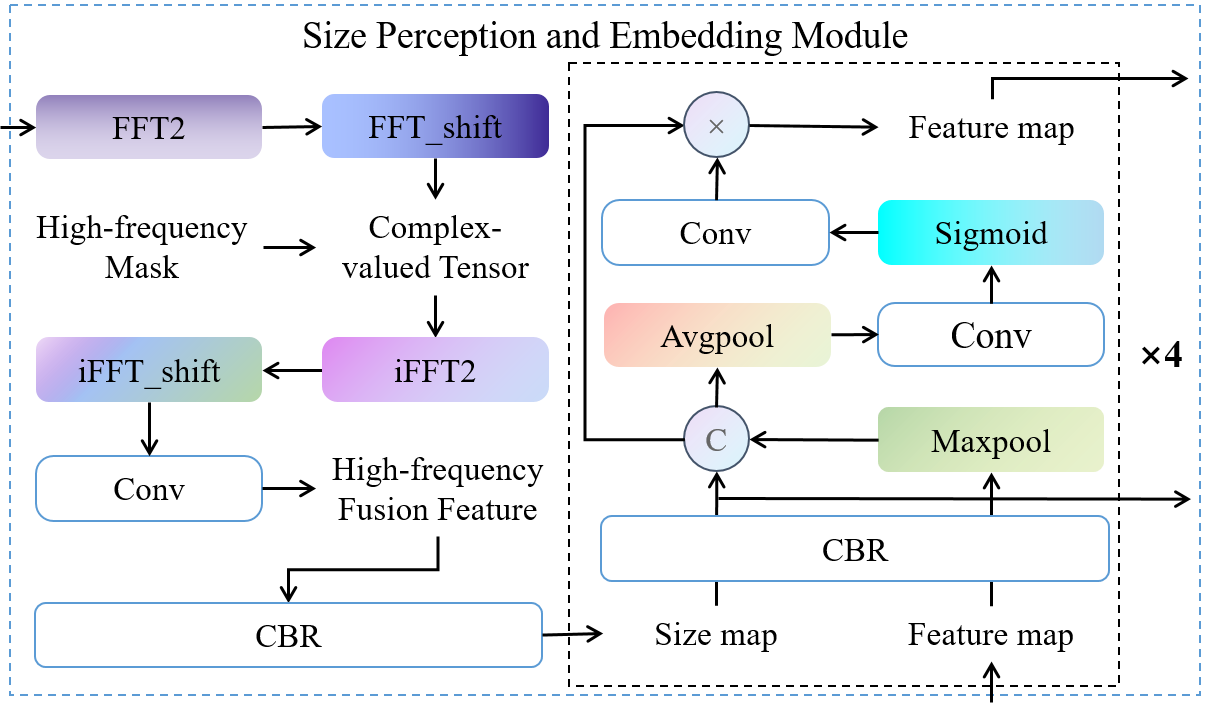}
\caption{SPE Module Architecture.}\label{fig6}
\end{figure}

Specifically, as shown in Figure \ref{fig6}, we use the lossless and efficient Fast Fourier Transform (FFT) \cite{FFT} to achieve this step. After receiving a batch of preprocessed images with shape $B \times 3 \times H \times W$, FFT is applied to convert them to the frequency domain, which is decomposed into frequency amplitude and phase components at the pixel level. Then, the central Low-frequency component is filtered, and the High-frequency component is inverted back to the spatial domain to obtain the High-frequency information of each channel. Finally, the channel dimension is adjusted by a layer of Conv to obtain the high-frequency prior information of the shape $B \times 1 \times H \times W$ that can characterize the size:
\begin{equation}
\begin{split}
f_{H}(i)=\left\{
\begin{aligned}
CBR(Conv_{1 \times 1}(FFT(f_{I}))), \quad i=0\\
CBR(f^{`}_{H}), \quad i>0\\
\end{aligned}
\right.
\label{eq5}
\end{split}
\end{equation}
where $i$ represents the Backbone network middle layer sequence number, $f_{I}$ represents the input to the APS network, $FFT$ represents the Fast Fourier Transform, $f^{`}_{H}$ represents the $f_{H}$ of the previous layer.

Secondly, how to embed the size information will directly affect the guidance effect of the Backbone network. Common fusion methods include operations such as summation, multiplication, channel concatenation, and spatial attention. However, summation and multiplication give equal importance to all spatial locations, causing the network to be susceptible to background noise or irrelevant regions. As mentioned before, spatial attention for fusion requires complex computational cost, and we can replace this process by passing the shallowly extracted prior information layer by layer. Therefore, we choose to concatenate the channel dimensions first, and then perform a simple channel attention mechanism to achieve the embedding of size information.

Specifically, the middle layer features of the backbone network will first go through a CBR layer to adjust the spatial dimension to align the scale. Then, concatenate with the size information in the channel dimension to obtain a feature map of $B \times (N+1) \times H \times W$. Then, simple channel attention is performed to amplify significant relationships between channels. The last Conv layer embeds the size information into the backbone network according to this relationship, resulting in the output of $B \times N \times H \times W$. In this way, it is a more efficient fusion process and can avoid the "blind" fusion by channel.

In summary, the SPE module can be expressed as follows:
\begin{equation}
\begin{split}
SPE(f_{H}, f_{B})=&Conv_{1 \times 1}(ATT_C(Concat_C(\\
&f_{H}, CBR(f_{B})))),
\label{eq6}
\end{split}
\end{equation}
where $ATT_C$ represents the channel attention, $Concat_C$ represents the concatenation operation of the channel dimension.

\subsection{ADD Structure and Loss Function}%\label{subsec4.2}

Learning fine-grained features on the basis of perceptual size information is still crucial for distinguishing morphologically similar species. Although hierarchical decoding structures can effectively help the network focus on fine-grained features through asynchronous learning across different levels \cite{xu2023fine, du_fine-grained_2020}, it is imprudent for traditional hierarchical decoders to handle size and fine-grained features in a unified manner in the context of ancient plant seed classification. To address this issue, we introduce an Asynchronous Decoupled Decoding (ADD) architecture with four training stages. ADD separates the spatial information from the original channel information, and this dual-view decoding strategy enables the network to learn discriminative seed features faster and more accurately.

Specifically, the first and second decoding phases of the ADD classification architecture use the traditional classification head \cite{ioffe2015batch} and the cross-entropy loss function \cite{rumelhart_learning_1986} to compute the supervised signal from a channel perspective:
\begin{equation}
\begin{split}
\mathcal{L}_{CE}=\frac{1}{B} \sum_{i} L_{i}=-\frac{1}{B} \sum_{i}^{B} \sum_{c=1}^{N} y_{i c} \log \left(p_{i c}\right),
\label{eq7}
\end{split}
\end{equation}
where $N$ is the number of classes, $y_{i c}$ is the sign function (0 or 1), and $p_{i c}$ is the predicted probability that an observation belongs to class $c$. Next we decouple in the third stage of four-asynchronous decoding, this is because the decoupling process should be performed after sufficiently learning the high-level semantic features of the image \cite{ge2021yolox, zhao2024detrs}. At this time, the feature coupling degree is the deepest and the decoupling effect is the best. It is worth noting that the fourth stage is not selected, because it only concatenates the channels of the features of the previous three stages to achieve the balancing operation, and there is no deeper expression in the high-level semantics.

\begin{figure}[t]
\centering
\includegraphics[width=0.95\linewidth]{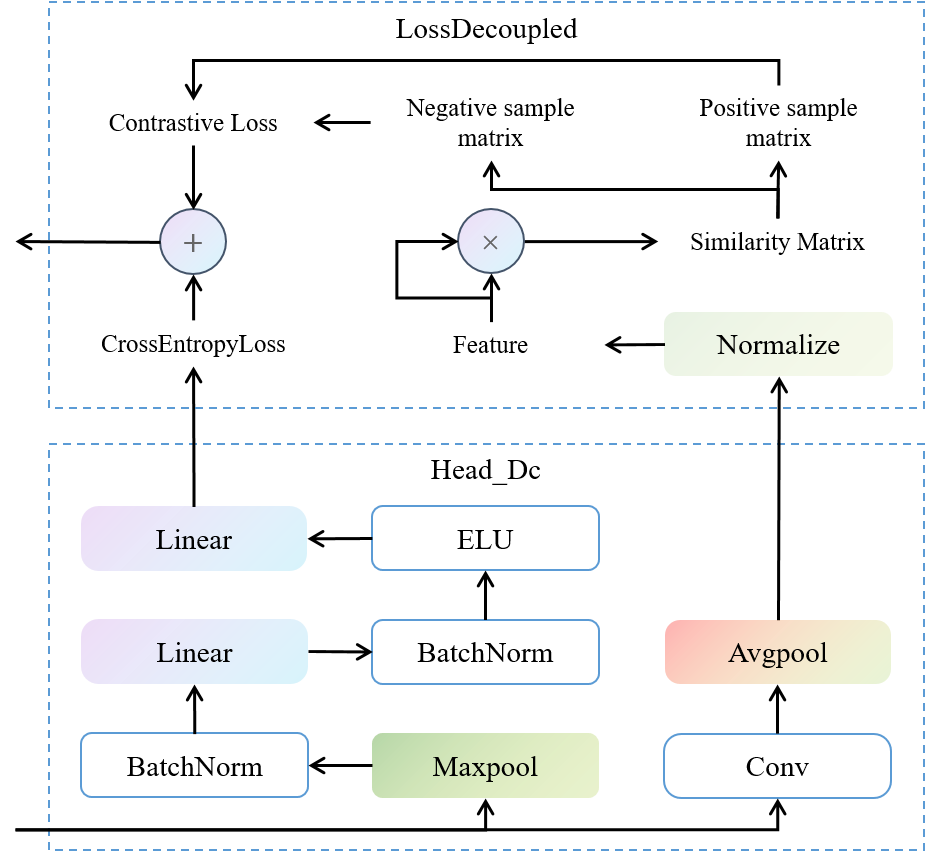}
\caption{Head\_Dc and LossComposite Architecture.}\label{fig7}
\end{figure}

As shown in Figure \ref{fig7}, we introduce Head\_Dc in the third stage. Head\_Dc first processes the large-scale feature maps output $\frac{H}{32} \times \frac{W}{32}$ from the Backbone by Max pooling to generate a batch of 1024-dimensional vectors, which are then passed through MLP to produce classification predictions in the channel direction. The cross-entropy loss is still used to generate supervision signals in the channel direction. However, the method of directly reducing the dimension of the spatial domain to obtain the channel result will lead to insufficient representation ability of spatial information such as size in the decoding stage. Therefore, the spatial branch preserves spatial information as an additional loss term by shrinking the channel information. The features of $\frac{H}{32} \times \frac{W}{32}$ are first projected into 1000 dimensional vectors for each sample to control the computational complexity. Then, for each sample, one channel dimension is selected from a vector of 1000 channel dimensions by MLP operation to represent the spatial information, resulting in the spatial representation. In addition, Cross-entropy loss focuses on dividing class boundaries and ignores the importance of intra-class compactness. We apply a Supervised Contrastive loss \cite{khosla_supervised_2020} to the generated spatial representation, which Narrows the same class and widens the different classes in spatial representation, in order to obtain better classification effect of ancient plant seeds. Specifically, after applying L2 normalization to project the vectors onto the unit sphere, the cosine similarity matrix is computed. Then, the corresponding penalty matrix is constructed to obtain the loss value:
\begin{equation}
\begin{split}
\mathcal{L}_{Contrast}=& \frac{1}{\mathfrak{B}^2} 
\sum_{i=1}^{\mathfrak{B}} \sum_{j=1}^{\mathfrak{B}}
[M_{ij}^{+} \left(1 - \cos(\mathbf{f}_i', \mathbf{f}_j') \right)
+ \\
&M_{ij}^{-} \, \max\left( \cos(\mathbf{f}_i', \mathbf{f}_j') - \kappa,\, 0 \right)],
\label{eq8}
\end{split}
\end{equation}
where $\mathbf{f}_i'= \frac{\mathbf{f}_i}{\|\mathbf{f}_i\|^2}$ represents the feature after L2 normalization, $\mathbf{f}_i \in \mathbb{R}^d$ represents the feature vector of the i-th seed sample, $M_{ij}^{+}$ represents the mask of the same class sample pair (positive sample pair: $y_{i}=y_{j}$), $M_{ij}^{-}$ represents the mask of the different class sample pair (negative sample pair: $y_{i}\neq y_{j}$), $y_{i}$ represents the label of the i-th sample, $\mathfrak{B}$ represents the total number of samples, $\kappa$ represents the heterogeneous sample penalty threshold. Therefore, Head\_Dc can be expressed as follows:
\begin{equation}
\begin{split}
Head\_Dc(f_{B})=&Head\_Tc(f_{B}), AP(\\ 
&Conv_{1 \times 1}(f_{B})), 
\label{eq8}
\end{split}
\end{equation}
where $Head\_Tc$ represents the traditional classification Head, the same as Head\_Con, Head\_1 and Head\_2, $AP$ represents average pooling, and $f_{B}$ represents the output of Backbone. The LossComposite combines the cross-entropy loss and the spatial distance loss to generate the final supervision signal from the perspective of dual domain:
\begin{equation}
\begin{split}
\mathcal{L}_{Composite}=\mathcal{L}_{CE}+\mathcal{L}_{Contrast}. 
\label{eq10}
\end{split}
\end{equation}

Finally, to balance the predictions across the three scales, APSNet adopts a standard multi-head structure.  Specifically, the fourth training stage concatenates the 3-scale feature maps along the channel dimension to generate the fused classification prediction, while combining the outputs of the four classification heads to produce the final prediction. Therefore, the total loss can be expressed as follows:
\begin{equation}
\begin{split}
\mathcal{L}^{}_{Train1}&=\mathcal{L}^{1}_{CE}.
\\
\mathcal{L}^{}_{Train2}&=\mathcal{L}^{2}_{CE}.
\\
\mathcal{L}^{}_{Train3}&=\mathcal{L}_{Composite}. 
\\
\mathcal{L}^{}_{Train4}&=\mathcal{L}^{Concat}_{CE}.
\label{eq11}
\end{split}
\end{equation}

\section{Experiment}%\label{sec5}

In this section, we present both quantitative and qualitative experiments. For the quantitative experiments, we conducted comprehensive comparisons and analyses of APSNet on the proposed APS dataset, and designed ablation studies to verify the effectiveness of the SPE and ADD. For the qualitative experiments, we demonstrate the significant improvements achieved by our method on the APS dataset.

\subsection{Implementation Details}

We implemented APSNet based on the well-established Pytorch framework. The experimental environment consists of Ubuntu 18.04, Python 3.12, Pytorch 2.6.0, CUDA 11.8, and CUDNN 9.0. To align with the practical conditions of archaeological experiments, model training and inference were conducted on an NVIDIA RTX 4090 GPU. For a fair comparison, all baseline methods were trained without using pretrained weights to reflect their actual performance. Unless otherwise specified, the experimental settings are as follows: input image size of 512 $\times$ 512, 200 training epochs with an early stopping mechanism set to 30 epochs, an initial learning rate of 1e-3, a weight decay of 5e-4, and a batch size of 16.

\subsection{Experiment Indicators}

In this section, we introduced the experimental indicators used.

\textbf{Accuracy:} Accuracy represents the proportion of the total number of correct predictions by the model.
\begin{equation}
\label{e5}
\text { Accuracy }=\frac{T P+T N}{T P+T N+F P+F N},
\end{equation}
where $TP$ denotes true positives, $TN$ denotes true negatives, $FP$ denotes false positives, and $FN$ denotes false negatives.

 %For uniform formatting, we changed the "P" in the body to italic.
\textbf{Precision:} Precision is the ratio of the number of correct tests to the number of positive tests.%  Equation~(\ref{e5}) is as follows:
\begin{equation}
\label{e6}
{ Precision }=\frac{T P}{T P+F P}.
\end{equation}

 %For uniform formatting, we changed the "R" in the body to italic.
 \textbf{Recall:} Recall is the ratio of the number of correct detections to the number of actual positive detections.%  Equation~(\ref{e6}) is as follows:
\begin{equation}
\label{e7}
{ Recall }=\frac{T P}{T P+F N}.
\end{equation}

 %For uniform formatting, we changed the "F1" in the body to italic.
\textbf{F1:} F1 is a measure of the accuracy of a model that takes into account both the accuracy and recall of a classification model.%  Equation~(\ref{e7}) is as follows:
\begin{equation}
\label{e8}
F 1 =\frac{2 \times { Precision } \times { Recall }}{ { Precision }+ { Recall }}.
\end{equation}

\subsection{Quantitative analysis}

In the quantitative analysis part, we design comparison experiments and ablation experiments to systematically evaluate the performance performance of the model under different Settings and verify the effectiveness of each module or strategy.

\subsubsection{Comparative Experiments}\label{subsubsec5.3.1}

% We first evaluated both state-of-the-art and classical methods on the APS dataset to establish a comprehensive benchmark, and subsequently conducted comparative experiments with APSNet. As shown in Table \ref{tab3}, we compared 28 classification methods, which can be broadly categorized into three groups. The results demonstrate that our proposed method achieves the best classification performance, surpassing the previous best method by 5.3\% in terms of Accuracy and by 12.1\% in terms of the F1 score. It can therefore be concluded that, in the task of ancient seed classification, the ability to perceive both inter-class size differences and fine-grained texture details is crucial. Existing methods lack explicit modeling of size-difference information, which ultimately limits their classification performance.

We first evaluated both state-of-the-art and classical methods on the APS dataset to establish a comprehensive benchmark, and subsequently conducted comparative experiments with APSNet. As shown in Table \ref{tab3}, we compared 28 classification methods, which can be broadly categorized into three groups.

Given the long-tail distribution of the APS dataset, the first group of classification methods evaluated two state-of the art classification methods designed for the long-tail problem, LTR (2022) and LOS (2025). Both approaches mitigate the imbalance issue by down-sampling classes with abundant samples to reduce the coverage gap for minority classes. However, our experimental results reveal that this strategy does not fundamentally resolve the imbalance in the APS dataset, as the recognition accuracy still tends to favor majority classes. Detailed Precision scores for each class are provided in Appendix \ref{tabB1}.

\begin{table}[h]
\centering
\caption{\centering Comparative Experiments on the APS Dataset}\label{tab3}
\resizebox{1\linewidth}{!}{%
\vbox{%
\begin{tabular}{@{}lccccc@{}}
\toprule
Method & Year & Accuracy  & Precision & Recall & F1 \\
\midrule
LTR & 2022 & 46.0\% & 37.0\% & 40.0\% & 31.0\% \\
LOS & 2025 & 58.2\% & 49.8\% & 55.2\% & 44.9\% \\
\hline
MetaFormer & 2022 & 57.9\% & 36.6\% & 34.2\% & 32.8\% \\
TransXNet & 2025 & 57.9\% & 50.4\% & 39.2\% & 37.0\% \\
Swin\_tiny & 2021 & 65.1\% & 39.5\% & 39.5\% & 37.9\% \\
2DMamba & 2025 & 61.2\% & 46.4\% & 42.9\% & 42.4\% \\
\makecell[l]{MobileViT\_\\small} & 2021 & 72.2\% & 47.1\% & 49.5\% & 44.9\% \\
CrossFormer++ & 2024 & 67.9\% & 48.2\% & 49.9\% & 47.8\% \\
ViT & 2020 & 68.3\% & 53.8\% & 51.7\% & 49.5\% \\
TranFG & 2022 & 75.5\% & 53.2\% & 56.3\% & 52.3\% \\
TCFormer & 2024 & 72.3\% & 55.7\% & 56.4\% & 52.7\% \\
SeaFormer & 2025 & 72.3\% & 54.3\% & 56.9\% & 53.2\% \\
IELT & 2023 & 76.6\% & 53.4\% & 61.0\% & 54.3\% \\
HAVT & 2023 & 75.3\% & 56.4\% & 58.4\% & 54.7\% \\
AA-Trans & 2023 & 77.5\% & 57.6\% & 57.5\% & 54.9\% \\
FET-FGVC & 2024 &  76.9\% & 57.2\% & 63.0\% & 56.7\% \\
\hline
ConvNext\_tiny & 2022 & 66.4\% & 44.8\% & 44.2\% & 43.7\% \\
\makecell[l]{MobileNetv3\_\\large} & 2019 & 76.0\% & 53.3\% & 54.4\% & 51.8\% \\
VGGNet11\_bn & 2014 & 75.0\% & 62.0\% & 58.4\% & 57.1\% \\
ResNest50 & 2020 & 81.2\% & 60.0\% & 61.9\% & 57.8\% \\
\makecell[l]{EfficientNetv2\_\\small} & 2021 & 79.3\% & 54.5\% & 61.6\% & 57.9\% \\
ResNet50 & 2015 & 82.0\% & 62.8\% & 64.6\% & 60.8\% \\
GoogleNet & 2014 & 79.2\% & 62.0\% & 63.5\% & 60.9\% \\
DenseNet201 & 2017 & 81.2\% & 61.5\% & 66.7\% & 61.5\% \\
GhostNetv3 & 2024 & 81.5\% & 63.3\% & 65.6\% & 61.5\% \\
\makecell[l]{ShuffleNetv2\_\\$\times$1.0} & 2018 & 80.9\% & 64.2\% & 65.4\% & 61.8\% \\
GhostNet & 2020 & 82.8\% & 68.6\% & 62.4\% & 62.0\% \\
GhostNetv2 & 2022 & 84.9\% & 66.3\% & 70.6\% & 65.4\% \\
\textbf{APSNet} & \textbf{Our} & \textbf{90.2\%} & \textbf{77.7\%} & \textbf{77.7\%} & \textbf{77.5\%} \\
\botrule
\end{tabular}}}

\footnotesize{Black bold represents the highest metric.}
% \footnotetext{Source: This is an example of table footnote. This is an example of table footnote.}
% \footnotetext[1]{Example for a first table footnote. This is an example of table footnote.}
% \footnotetext[2]{Example for a second table footnote. This is an example of table footnote.}
\end{table}

The second group of classification methods, published between 2020 and 2025, all rely on pre-trained weights from large-scale general-purpose datasets such as ImageNet-1K to maintain accuracy. Most of these models are based on Transformer architectures, which are often large and computationally complex. Among them, the fine-grained classification model FET-FGVC achieves the highest F1 score on the APS dataset, delivering the best overall performance within this group. The third group consists of classical image classification methods, typically designed with CNN-based architectures. Compared to Transformer-based models, these approaches have lower complexity and stronger feature extraction capabilities, and they outperform methods without pre-trained weights. Within this group, we selected variants of comparable scale to ResNet50 for evaluation and comparison.

\begin{table}[h]
\caption{\centering Ablation experiments of prior information in the SPE module}\label{tab4}
\centering
\resizebox{1\linewidth}{!}{
\vbox{
\begin{tabular}{@{}lccccc@{}}
\toprule
\makecell[l]{Method\\ (APSNet)} & Entropy & Accuracy  & Precision & Recall & F1 \\
\midrule
+Edge & 0.1 & 87.0\% & 75.0\% & 73.8\% & 74.4\% \\
+Mask & 0.3 & 88.2\% & 76.0\% & \textbf{78.9\%} & \textbf{77.5\%} \\
+Sobel & 0.6 & 88.4\% & 75.0\% & 72.9\% & 73.9\% \\
+Raw & 0.9 & 88.8\% & 76.1\% & 77.2\% & 76.6\% \\
+LF & 1.3 & 89.2\% & \textbf{78.3\%} & 76.2\% & 77.2\% \\
\textbf{+HF (our)} & \textbf{1.4} & \textbf{90.2\%} & 77.7\% & 77.7\% & \textbf{77.5\%} \\
\botrule
\end{tabular}}}

\footnotesize{Black bold represents the metric improvement.}
% \footnotetext{Source: This is an example of table footnote. This is an example of table footnote.}
% \footnotetext[1]{Example for a first table footnote. This is an example of table footnote.}
% \footnotetext[2]{Example for a second table footnote. This is an example of table footnote.}
\end{table}

The results demonstrate that our proposed method achieves the best classification performance, surpassing the previous best method by 5.3\% in terms of Accuracy and by 12.1\% in terms of the F1 score. It can therefore be concluded that, in the task of ancient seed classification, the ability to perceive both inter-class size and fine-grained texture details is crucial. Existing methods lack explicit modeling of size information, which ultimately limits their classification performance. Moreover, the results justify our choice of ResNet50 as the Backbone of APSNet. Specifically, we selected the Backbone based on the most critical metric in classification tasks—Accuracy. Although GhostNet achieves the best Accuracy score, it was originally designed for edge devices, resulting in very low GPU utilization. In contrast, given that our archaeological experimental environment employs high-performance equipment, APSNet uses ResNet50 as the Backbone, and its GPU utilization far exceeds that of the GhostNet series. 

\subsubsection{Ablation Experiments}\label{subsubsec5.3.2}

To validate the effectiveness of the proposed SPE module and ADD structure, we designed ablation experiments.    Among them, we additionally analyze the effect of different prior information representation sizes for the SPE module.    For the ADD architecture, we focus on ablating the newly introduced Head\_Dc, because phased training and multi-head architecture are proven methods for fine-grained and multi-scale feature extraction. Its effectiveness has been demonstrated in previous work \cite{rusu2016progressive, tan2021efficientnetv2}.    Since the Backbone of APSNet is ResNet50, we evaluated the performance of the SPE module and Head\_Dc on four CNN-based architectures: GoogleNet, EfficientNet, DenseNet, and ShuffleNet.    In addition, to broaden the scope and persuasiveness of the experiments, we applied Head\_Dc to ViT-based methods to test whether dual-domain voting could improve accuracy on the APS dataset.    We did not apply ablation tests of the SPE module to ViT, as SPE module is specifically designed for CNN architectures.

\begin{table}[t]
\caption{\centering Ablation Experiments of the SPE Module and Head\_Dc}\label{tab5}
\centering
\resizebox{1\linewidth}{!}{
\vbox{
\begin{tabular}{@{}lcccc@{}}
\toprule
Method & Accuracy  & Precision & Recall & F1 \\
\midrule
ViT  & 68.3\% & 53.8\% & 51.7\% & 49.5\% \\
\textbf{ViT+Head\_Dc} & \textbf{75.5\%} & \textbf{56.6\%} & \textbf{55.1\%} & \textbf{54.0\%} \\
\hline
EfficientNetv2\_small & 79.3\% & 54.5\% & 61.6\% & 57.9\% \\
\textbf{\makecell[l]{EfficientNetv2\_small \\+SPE}} & \textbf{78.6\%} & \textbf{63.7\%} & \textbf{66.3\%} & \textbf{59.6\%} \\
\textbf{\makecell[l]{EfficientNetv2\_small \\+Head\_Dc}} & \textbf{80.6\%} & \textbf{60.5\%} & \textbf{64.1\%} & \textbf{59.6\%} \\
\hline
GoogleNet & 79.2\% & 62.0\% & 63.5\% & 60.9\% \\
\textbf{GoogleNet+SPE} & \textbf{81.3\%} & \textbf{63.5\%} & \textbf{65.7\%} & \textbf{62.9\%} \\
\textbf{GoogleNet+Head\_Dc} & \textbf{82.2\%} & \textbf{64.3\%} & \textbf{63.9\%} & \textbf{63.0\%} \\
\hline
DenseNet201 & 81.2\% & 61.5\% & 66.7\% & 61.5\% \\
\textbf{DenseNet201+SPE} & \textbf{84.7\%} & \textbf{68.9\%} & \textbf{68.6\%} & \textbf{65.5\%} \\
\textbf{DenseNet201+Head\_Dc} & \textbf{85.5\%} & \textbf{73.0\%} & \textbf{71.3\%} & \textbf{68.9\%} \\
\hline
ShuffleNetv2\_$\times$1.0 & 80.9\% & 64.2\% & 65.4\% & 61.8\% \\
\textbf{\makecell[l]{ShuffleNetv2\_$\times$1.0 \\+SPE}} & \textbf{84.1\%} & \textbf{66.5\%} & 64.2\% & 61.5\% \\
\textbf{\makecell[l]{ShuffleNetv2\_$\times$1.0 \\+Head\_Dc}} & \textbf{81.4\%} & \textbf{64.6\%} & \textbf{67.8\%} & 61.2\% \\
\botrule
\end{tabular}}}

\footnotesize{Black bold represents the metric improvement.}
% \footnotetext{Source: This is an example of table footnote. This is an example of table footnote.}
% \footnotetext[1]{Example for a first table footnote. This is an example of table footnote.}
% \footnotetext[2]{Example for a second table footnote. This is an example of table footnote.}
\end{table}

Table \ref{tab4} shows our classification performance using Original Image (Raw), Edge, Mask, Sobel, Low-frequency (LF) and High-frequency (HF) as prior information representing the size. Among them, the Mask is obtained through the point prompt of SAM \cite{kirillov2023segment}, and the Edge is obtained by filtering the Mask. The results prove our analysis in Section 4.1, and the high-frequency components achieve the best results in guiding the Backbone network to learn the seed size due to their rich information representation. Moreover, the performance ranking of the most important Accuracy metrics is consistent with the ranking of the average entropy value, suggesting that the information richness perspective to measure whether prior information effectively characterizes size is correct. Surprisingly, Mask achieves excellent performance on Recall and F1 metrics. We analyze this because its strict division of foreground and background has an implicit advantage in characterizing the seed size. Therefore, we do further visual analysis in Section \ref{subsubsec5.4.2}.

\begin{figure*}[h]
\centering
\includegraphics[width=1\textwidth]{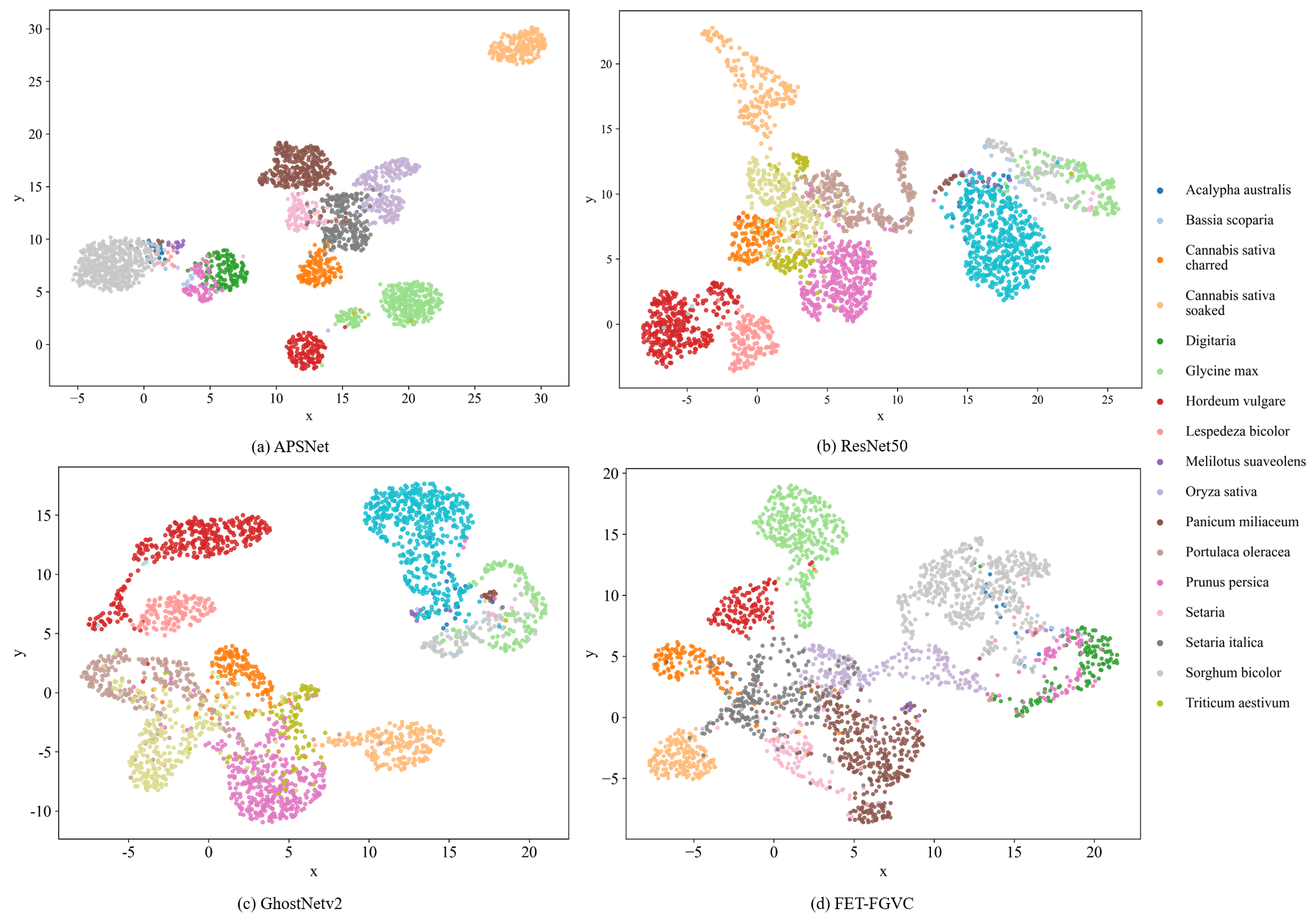}
\caption{Visualization of classification ability on the test set. Effect of APSNet, baseline models ResNet50, GhostNetv2, and FET-FGVC method on UMAP.}\label{fig8}
\end{figure*}

Table \ref{tab5} presents the ablation experiments for the two proposed components. For the SPE module, we observe that all four CNN-based methods achieve improvements after incorporating size information. Specifically, EfficientNetV2 achieves the largest gain in Precision (+9.2\%), GoogleNet in Recall (+2.2\%), DenseNet201 in Precision (+7.4\%), and ShuffleNetV2 in Accuracy (+3.2\%).  These results show that it is crucial to consider both size and texture detail information in ancient seed classification, which verifies the effectiveness of the SPE module.

Among the five classification methods, the ensemble of Head\_Dc shows a clear improvement in all most metrics.   Specifically, the highest Accuracy gain is observed with ViT (+7.2\%), while DenseNet201 achieves the largest improvements in Precision (+11.5\%), Recall (+4.6\%), and F1 score (+7.4\%).   These results provide strong evidence for the importance of decoupling spatial information in ancient seed classification tasks, and further demonstrate the reliability and effectiveness of the Head\_Dc design.

\subsection{Qualitative analysis}

In the qualitative analysis part, we carry out the visualization experiment of classification effect and feature extraction, in order to more intuitively show the discrimination ability of the model on different categories of samples, and explore the learning and discrimination effect of the model on the target in the feature space, so as to verify the rationality and interpretability of its internal representation.

\subsubsection{Classification Effect Comparison}

We compared APSNet with three other models using UMAP \cite{mcinnes2018umap} visualization: the baseline ResNet50, GhostNetv2 (the best-performing CNN-based model), and FET-FGVC (the best-performing Transformer-based model). For fairness, all four models used the best-trained checkpoints on the APS dataset, with identical UMAP parameters. 

As shown in Figure \ref{fig8}, APSNet demonstrates the strongest classification ability, characterized by tighter intra-class clusters and larger inter-class separations. This highlights the critical role of automatically perceiving size information in ancient seed classification. Furthermore, ResNet50 shows little improvement after training compared to its pre-trained state (Figure \ref{fig4.2}(c)), revealing the limitations of traditional methods that fail to model size and further validating the effectiveness of our proposed approach.

\subsubsection{Feature visualization of different prior information}\label{subsubsec5.4.2}

\begin{figure}[h]
\centering
\includegraphics[width=1\linewidth]{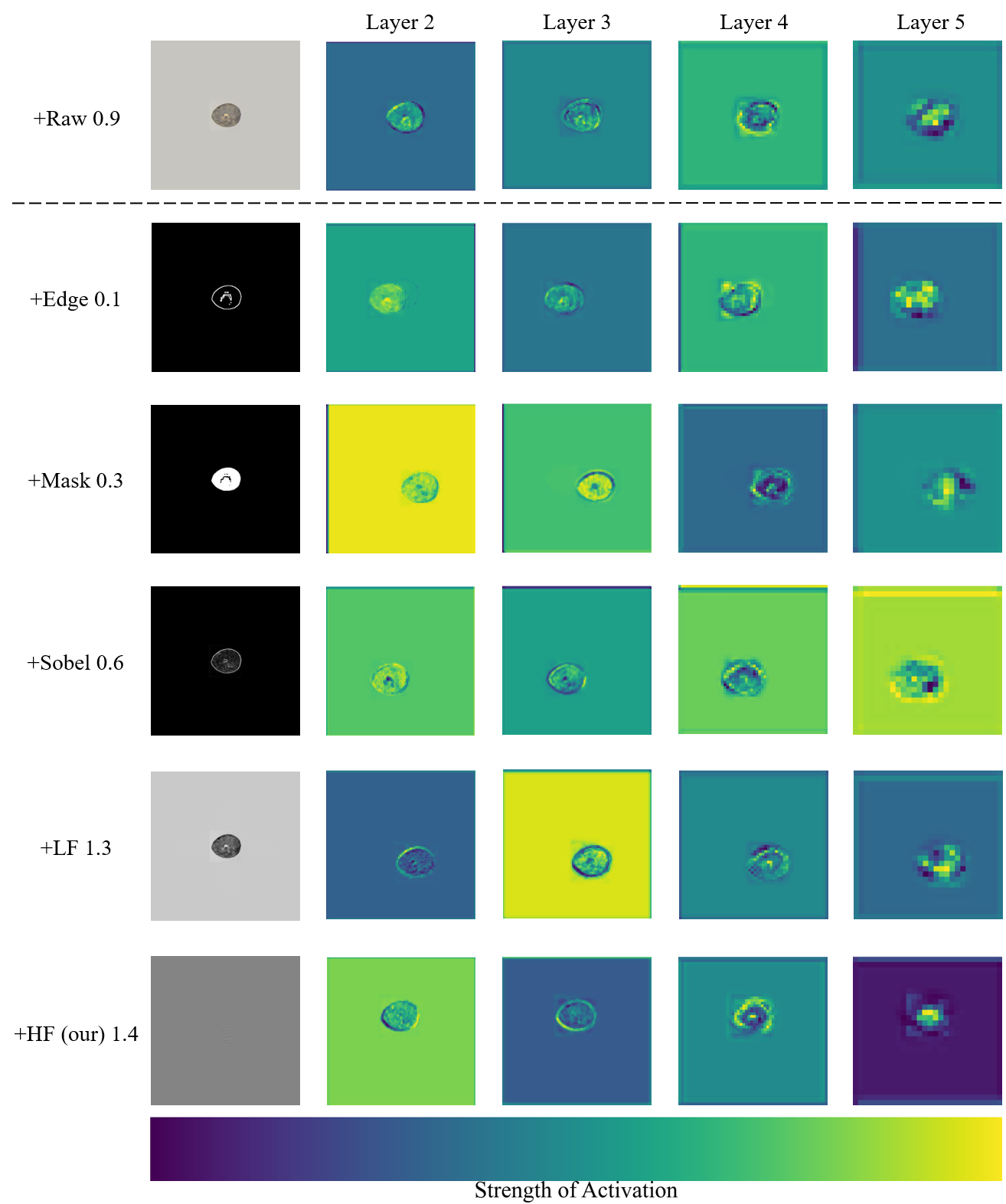}
\caption{The feature transfer variation of different prior information in the SPE module.}\label{fig9}
%The red area indicates the region that the method focuses on, which can be regarded as the discriminative feature basis for method classification.}\label{fig9}
\end{figure}

Based on the information richness, we further analyze the change of features from deep to shallow layers guided by different prior information in Backbone network to evaluate their respective effectiveness (taking \textit{Acalypha australis} as an example). Specifically, we visualize the characteristics of the SPE modules in APSNet docking layers 2, 3, 4, and 5 of the Backbone network. The activation strength of the feature is obtained by averaging over the channel layers, with blue indicating weak activation and yellow indicating strong activation. As shown in Figure \ref{fig9}, only FS and Mask guide the Backbone network to diverge from the global to the focused foreground among the six cases with different representation sizes of prior information. The FS size information allows Backbone to focus on the foreground faster than Mask. This validates our analysis in Sections \ref{subsec4.1} and \ref{subsubsec5.3.2}, showing that FS has rationality as size information for prior bootstrapping.

\subsubsection{Feature Heatmap Comparison}

To examine the improvements of the proposed APSNet in feature processing, we visualized attention heatmaps on the APS test set using Grad-CAM \cite{selvaraju2017grad}. Specifically, we extracted the final Conv feature maps from four methods and projected their attention onto the original images. Following the setup in Section \ref{subsubsec5.3.1}, we selected ResNet50 as the baseline, GhostNetv2 as the representative CNN-based model, and FET-FGVC as the representative Transformer-based model.

\begin{figure}[h]
\centering
\includegraphics[width=0.85\linewidth]{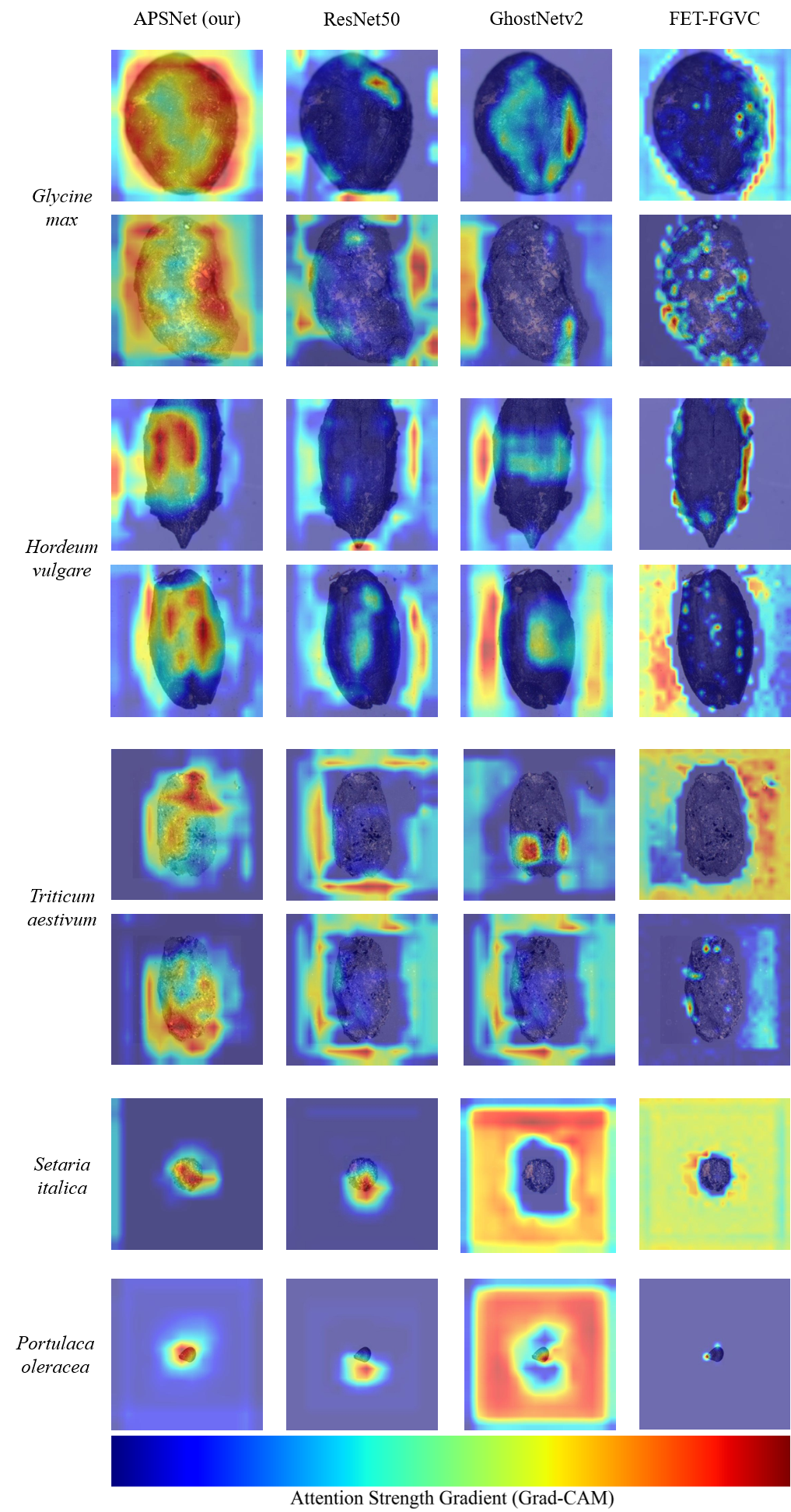}
\caption{Visualization of the attention heatmap on the test set.}\label{fig10}
\end{figure}

As shown in Figure \ref{fig10}, we selected five categories for visualization, including two large-scale targets (\textit{Glycine max}, \textit{Hordeum vulgare}), one medium-scale target (\textit{Triticum aestivum}), and two small-scale targets (\textit{Setaria italica}, \textit{Portulaca oleracea}). Among them, \textit{Portulaca oleracea} belongs to the few-shot class. The results indicate that APSNet achieves significant improvements over the baseline ResNet50, particularly for large- and medium-scale targets. This demonstrates that both the SPE module and ADD classification structure  positively contribute to enhancing the baseline model’s performance.

In the case of GhostNetv2, its dilated convolution design indeed demonstrates the advantage of an enlarged receptive field, showing clear improvements in contour extraction for medium- and small-scale targets. However, it performs less effectively on large-scale targets. Moreover, due to the lack of explicit modeling of size information, GhostNetv2 fails to accurately perceive contour information. In addition, its large receptive field design limits the network’s focus on fine-grained texture details.

As a representative fine-grained classification method, FET-FGVC demonstrates strong capability in processing detailed information, albeit at the cost of being computationally heavy and complex. In contrast, the more lightweight APSNet shows superior performance in contour extraction. For instance, in the categories \textit{Glycine max} and \textit{Triticum aestivum}, APSNet outperforms FET-FGVC in capturing contour features. When it comes to texture detail processing, FET-FGVC surpasses APSNet in certain aspects—for example, in the \textit{Hordeum vulgare} class. However, APSNet compensates by effectively capturing both the two longitudinal ridges along the ventral groove of \textit{Hordeum vulgare} and inter-class size information. For small-scale categories such as \textit{Triticum aestivum} and \textit{Portulaca oleracea}, APSNet demonstrates the ability to extract size cues, leveraging them as discriminative features for accurate classification.

These results collectively demonstrate the superiority of APSNet over existing methods in the task of ancient seed classification. Moreover, the relatively poor performance of current approaches on the APS dataset underscores the necessity of establishing this dataset, which serves as a foundational reference for advancing research in this field.

\section{Conclusion}%\label{sec13}

In this study, we proposed and systematically validated the APS dataset and the APSNet framework for the recognition of ancient charred plant seeds.       The APS dataset spans a broad temporal and spatial distribution of representative archaeological samples, addressing the long-standing lack of standardized data resources in archaeobotany.       To tackle the challenges of high inter-class similarity and large intra-class variation, we designed the APSNet architecture, which can automatically perceive size informaton.       Specifically, the SPE module and the Head\_Dc-based ADD structure effectively enhance the model’s ability to capture both size and texture details.       Experimental results demonstrate that APSNet significantly outperforms existing methods in terms of both accuracy and robustness.       Furthermore, Grad-CAM and UMAP visualizations provide additional evidence of its advantages in capturing size and texture features. However, the current work has not solved the problem of long-tail distribution of dataset categories.       In future research, we plan to explore effective approaches for handling such cases.       We hope that this study can provide new perspectives and practical foundations for the digitalization and intelligent advancement of archaeobotanical research.

\section{Declarations}%\label{sec13}

\bmhead{Data availability}

The proposed method is publicly available on \href{https://github.com/RuiXing123/APSNet}{Github}, and the ASP dataset is available from the corresponding author upon reasonable request.

\bmhead{Acknowledgements}

We would like to express our gratitude for the researchers who carried out the archaeological work and performed the image data acquisition.

\bmhead{Funding Information}

This work was supported by Key Scientific Research Base of Paleoenvironment Reconstruction and Subsistence Studies in Archaeology (Shandong University), National Cultural Heritage Administration (Project No. SDULA202504).

\bmhead{Authors' contributions}

R.X. wrote the main manuscript text. R.C., Y.W., C.W., Z.T., F.W., H.W. and S.K. reviewed the manuscript.

\bmhead{Competing interests}

The authors declare that they have no competing interests.

\begin{appendices}

\section{Classification Criteria}\label{secA1}

The APS dataset categories are strictly defined according to archaeobotanical standards for ancient seeds (Table \ref{tabA1}). During data collection, samples meeting the definitions were retained, while those that did not were treated as noise and filtered out.
\begin{table*}[h]
\caption{\centering Classification Criteria of the APS Dataset}\label{tabA1}
\centering
\begin{tabular}{@{}lc@{}}
\toprule
Classes & Criteria \\
\midrule
\textit{Acalypha australis} & \makecell[c]{The seed is obovoid, with a rounded apex, slightly pointed\\ base, and hilum at the basal inner side.} \\
\hline
\textit{Bassia scoparia} & \makecell[c]{The seed is obovoid, with a pointed apex, a rounded base,\\ and a horseshoe-shaped embryo.} \\
\hline
\textit{Cannabis sativa charred} & \multirow{2}{*}{\makecell[c]{The seed is compressed-ovoid, convex on both sides, with\\ a small apical protrusion, rounded base, and marginal ridges.}} \\
\textit{Cannabis sativa soaked} &  \\
 \hline
\textit{Digitaria} & \makecell[c]{The caryopsis is lanceolate, with a rounded dorsal protuberance,\\ flat ventral side, blunt apex, slightly pointed base, distinct\\ embryo over one-third of its length, and a small elliptical hilum.} \\
\hline
\textit{Glycine max} & \makecell[c]{The hilum is near the ventral center, narrowly elliptic to\\ rectangular-elliptic;  the embryo is curved, and the transverse\\ section is elliptic.} \\
\hline
\textit{Hordeum vulgare} & \makecell[c]{The caryopsis is elliptical, dorsally and ventrally convex, widest\\ and thickest at the center. The dorsal surface has five faint\\ ridges; the ventral groove widens toward the apex, about\\ one-fifth the width and half the thickness, separating into a\\ “V” shape at the top one-tenth. The embryo occupies one-third\\ of the length, with an elliptical transverse section.} \\
\hline
\textit{Lespedeza bicolor} & \makecell[c]{The seed is reniform-ovoid, radicle shorter than cotyledons, dorsal\\ side blunt, ventral side obliquely truncated; hilum circular with\\ protruding margin, no endosperm.} \\
\hline
\textit{Melilotus suaveolens} & \makecell[c]{The seed is ovoid-elliptical, smooth, with a blunt apex\\ and obliquely concave base. The radicle is shorter than the\\ cotyledons, and a small amount of endosperm is present.} \\
\hline
\textit{Oryza sativa} & \makecell[c]{The caryopsis is elliptic to oblong-rectangular, slightly laterally\\ compressed, with two longitudinal ridges. The lateral embryo is\\ about one-fourth the length and half the width; transverse section\\ is elliptical.} \\
\hline
\textit{Portulaca oleracea} & \makecell[c]{The seed is dark brown, less than 1 mm in diameter, with small\\ tubercles on the surface.} \\
\hline
\textit{Prunus persica} & \makecell[c]{The seed is elliptical, slightly flattened, biconvex-lens shaped,\\ with a small apical tip and fine longitudinal grooves. The\\ hilum is centrally basal, narrowly elliptical, and inconspicuous;\\ transverse section is elliptical.} \\
\hline
\textit{Setaria} & \multirow{3}{*}{\makecell[c]{The caryopsis is broadly elliptical, with a bluntly pointed apex,\\ a convex dorsal side, a flat ventral side, and an elliptical embryo.}} \\
\textit{Panicum miliaceum} &  \\
\textit{Setaria italica} &  \\
\hline
\textit{Sorghum bicolor} & \makecell[c]{The caryopsis is elliptical, slightly convex on both sides, with\\ the apex slightly protruding.} \\
\hline
\textit{Triticum aestivum} & \makecell[c]{The caryopsis is ovate-elliptical, with a convex dorsal side\\ and flat ventral side; the embryo is nearly circular, one-third\\ the fruit length, and the transverse section is circular.} \\
\botrule
\end{tabular}
% \footnotetext{Source: This is an example of table footnote. This is an example of table footnote.}
% \footnotetext[1]{Example for a first table footnote. This is an example of table footnote.}
% \footnotetext[2]{Example for a second table footnote. This is an example of table footnote.}
\end{table*}

\section{Class-level Comparative Experiments}\label{secA2}

Table \ref{tabB1} presents the Precision metric for each class in the comparative experiments, showing that APSNet achieves high precision across most categories, particularly in those with larger sample sizes. Specifically, APSNet achieves the highest Precision metric in eight categories. While the other methods  although will outperform our method on some categories, they are extremely imbalanced. For example, GhostNet achieves 100\% Precision on the \textit{Bassia scoparia} class. This phenomenon does not indicate that GhostNet is superior to our proposed method. Because, the Precision on \textit{Acalypha australis} and \textit{Prunus persica} categories is 0\%, and on other categories is much lower than our method.
\begin{table*}[h]
\caption{\centering Class-level Comparative Experiments (Precision Metric)}\label{tabB1}
\centering
\begin{tabular}{@{}lcccccc@{}}
\toprule
Method & \textit{\makecell[c]{Acalypha\\ australis}}	& \textit{\makecell[c]{Bassia\\ scoparia}} & \textit{\makecell[c]{Cannabis\\ sativa charred}} & \textit{\makecell[c]{Cannabis\\ sativa soaked}} & \textit{Digitaria} \\
\midrule
LTR & 1\% & 7\% & 57\% & 79\% & 33\% \\
LOS & 18\% & 8\% & 59\% & 80\% & 77\% \\
\hline
MetaFormer & 0\% & 0\% & 45\% & 68\% & 58\% \\
TransXNet & 4\% & 0\% & 100\% & 96\% & 52\% \\
Swintiny & 0\% & 0\% & 67\% & 80\% & 53\% \\
2DMamba & 0\% & 0\% & 46\% & 97\% & 61\% \\
MobileViTsmall & 0\% & 0\% & 74\% & 70\% & 63\% \\
CrossFormer++ & 0\% & 0\% & 68\% & 96\% & 72\% \\
ViT & 0\% & 43\% & 97\% & 97\% & 54\% \\
TranFG & 8\% & 25\% & 84\% & 94\% & 33\% \\
TCFormer & 17\% & 33\% & 96\% & 87\% & 76\% \\
SeaFormer & 0\% & 25\% & 65\% & 81\% & 18\% \\
IELT & 18\% & 31\% & 69\% & 98\% & 0\% \\
HAVT & 10\% & 22\% & 85\% & 95\% & 66\% \\
AA-Trans & 23\% & 13\% & 87\% & 96\% & 81\% \\
FET-FGVC & 4\% & 21\% & 92\% & 91\% & 62\% \\
\hline
ConvNext\_tiny & 0\% & 0\% & 62\% & 85\% & 67\% \\
MobileNetv3\_large & 0\% & 11\% & 65\% & 97\% & 66\% \\
VGGNet11\_bn & 33\% & 63\% & 77\% & 88\% & 83\% \\
ResNest50 & 0\% & 50\% & 96\% & 89\% & 72\% \\
EfficientNetv2\_small & 18\% & 7\% & 95\% & 81\% & 80\% \\
ResNet50 & 10\% & 50\% & 80\% & 95\% & 78\% \\
GoogleNet & 13\% & 46\% & 95\% & 99\% & 49\% \\
DenseNet201 & 0\% & 20\% & 88\% & 92\% & 84\% \\
GhostNetv3 & 28\% & 33\% & 88\% & 94\% & 84\% \\
ShuffleNetv2\_$\times$1.0 & 40\% & 42\% & 64\% & 99\% & 78\% \\
GhostNet & 0\% & \textbf{100\%} & \textbf{100\%} & 94\% & \textbf{88\%} \\
GhostNetv2 & \textbf{60\%} & 15\% & 86\% & 98\% & 83\% \\
\textbf{APSNet} & 35\% & 24\% & 96\% & \textbf{100\%} & 70\% \\
\botrule
\end{tabular}

\footnotesize{Continued on next page.}
% \footnotetext[1]{Example for a first table footnote. This is an example of table footnote.}
% \footnotetext[2]{Example for a second table footnote. This is an example of table footnote.}
\end{table*}

\begin{table*}[h]
\caption*{\centering \textbf{Continued Table} \ref{tabB1}: Class-level Comparative Experiments (Precision Metric)}\label{tabB2}
\centering
\begin{tabular}{@{}lcccccc@{}}
\toprule
Method & \textit{\makecell[c]{Glycine\\ max}} & \textit{\makecell[c]{Hordeum\\ vulgare}} & \textit{\makecell[c]{Lespedeza\\ bicolor}} & \textit{\makecell[c]{Melilotus\\ suaveolens}} & \textit{\makecell[c]{Setaria\\ italica}} & \textit{\makecell[c]{Oryza\\ sativa}} \\
\midrule
LTR & \textbf{100\%} & 69\% & 6\% & 23\% & 95\% & 31\%\\
LOS & 94\% & 91\% & 3\% & 26\% & 61\% & 56\%\\
\hline
MetaFormer & 97\% & 93\% & 0\% & 0\% & 58\% & 30\%\\
TransXNet & \textbf{100\%} & 83\% & 14\% & 75\% & 87\% & 51\%\\
Swintiny & 98\% & 93\% & 21\% & 0\% & 75\% & 38\%\\
2DMamba & 98\% & 87\% & 50\% & 60\% & 61\% & 33\%\\
MobileViTsmall & 90\% & 91\% & 17\% & 12\% & 95\% & 66\%\\
CrossFormer++ & 99\% & 95\% & 23\% & 56\% & 86\% & 36\%\\
ViT & \textbf{100\%} & 74\% & 18\% & 79\% & 84\% & 29\%\\
TranFG & 97\% & 87\% & 16\% & 36\% & 91\% & 66\%\\
TCFormer & 99\% & 96\% & 35\% & 9\% & 90\% & 66\%\\
SeaFormer & 99\% & 92\% & 16\% & 44\% & 95\% & 45\%\\
IELT & 97\% & 89\% & 6\% & 73\% & 86\% & 71\%\\
HAVT & 99\% & 77\% & 13\% & 50\% & 81\% & 56\%\\
AA-Trans & 98\% & 86\% & 13\% & 46\% & 89\% & 68\%\\
FET-FGVC & 98\% &  97\% & 19\% & 50\% & 94\% & 89\%\\
\hline
ConvNext\_tiny & 97\% & 94\% & 20\% & 83\% & 88\% & 49\%\\
MobileNetv3\_large & 88\% & 98\% & 15\% & 72\% & 94\% & 82\%\\
VGGNet11\_bn & 97\% & 88\% & 26\% & 47\% & 95\% & 75\%\\
ResNest50 & 95\% & 81\% & 24\% & 56\% & 96\% & 79\%\\
EfficientNetv2\_small & 99\% & 91\% & 21\% & 76\% & 98\% & 94\%\\
ResNet50 & 89\% & 87\% & 26\% & 67\% & 95\% & 81\%\\
GoogleNet & 98\% & 90\% & 31\% & 45\% & 96\% & 64\%\\
DenseNet201 & 95\% & 95\% & 21\% & 67\% & 84\% & 70\%\\
GhostNetv3 & 99\% & 92\% &11\% & 74\% & \textbf{98\%} & 69\%\\
ShuffleNetv2\_$\times$1.0 & 83\% & 98\% & 24\% & 70\% & 96\% & 70\%\\
GhostNet & 99\% & 97\% & 50\% & \textbf{93\%} & 95\% & 78\%\\
GhostNetv2 & 99\% & 95\% & 24\% & 72\% & \textbf{98\%} & 84\%\\
\textbf{APSNet} & 96\% & \textbf{99\%} & \textbf{69\%} & \textbf{93\%} & \textbf{98\%} & \textbf{100\%}\\
\botrule
\end{tabular}

\footnotesize{Continued on next page.}
% \footnotetext[1]{Example for a first table footnote. This is an example of table footnote.}
% \footnotetext[2]{Example for a second table footnote. This is an example of table footnote.}
\end{table*}

\begin{table*}[h]
\caption*{\centering \textbf{Continued Table} \ref{tabB1}: Class-level Comparative Experiments (Precision Metric)}\label{tabB3}
\centering
\begin{tabular}{@{}lcccccc@{}}
\toprule
Method & \textit{\makecell[c]{Panicum\\ miliaceum}} & \textit{\makecell[c]{Prunus\\ persica}} & \textit{\makecell[c]{Portulaca\\ oleracea}} & \textit{Setaria} & \textit{\makecell[c]{Sorghum\\ bicolor}} & \textit{\makecell[c]{Triticum\\ aestivum}} \\
\midrule
LTR & 30\% & 0\% & 0\% & 2\% & 40\% & 50\%\\
LOS & 78\% & 20\% & 8\% & 29\% & 55\% & 85\%\\
\hline
MetaFormer & 25\% & 0\% & 0\% & 28\% & \textbf{89\%} & 28\%\\
TransXNet & 83\% & 0\% & 0\% & 30\% & 14\% & 67\%\\
Swintiny & 64\% & 0\% & 0\% & 25\% & 24\% & 33\%\\
2DMamba & 35\% & 0\% & 0\% & 40\% & 81\% & 38\%\\
MobileViTsmall & 71\% & 0\% & 0\% & 56\% & 41\% & 54\%\\
CrossFormer++ & 50\% & 0\% & 0\% & 36\% & 54\% & 45\%\\
ViT & 75\% & 7\% & 6\% & 58\% & 32\% & 64\%\\
TranFG & 89\% & 0\% & 2\% & 37\% & 74\% & 65\%\\
TCFormer & 55\% & 0\% & 33\% & 41\% & 63\% & 51\%\\
SeaFormer & 73\% & 0\% & 67\% & 58\% & 38\% & 47\%\\
IELT & 88\% & 5\% & 0\% & 36\% & 63\% & 77\%\\
HAVT & \textbf{95\%} & 0\% & 26\% & 36\% & 77\% & 69\%\\
AA-Trans & 91\% & 0\% & 6\% & 35\% & 83\% & 65\%\\
FET-FGVC & 89\% & 3\% & 5\% & 37\% & 74\% & 78\%\\
\hline
ConvNext\_tiny & 52\% & 0\% & 0\% & 40\% & 45\% & 24\%\\
MobileNetv3\_large & 71\% & 2\% & 0\% & 42\% & 47\% & 56\%\\
VGGNet11\_bn & 50\% & 20\% & 20\% & 63\% & 54\% & 76\%\\
ResNest50 & 81\% & 5\% & 15\% & 44\% & 45\% & 91\%\\
EfficientNetv2\_small & 68\% & 3\% & 0\% & 47\% & 37\% & 92\%\\
ResNet50 & 79\% & 8\% & 25\% & 50\% & 62\% & 86\%\\
GoogleNet & 72\% & 0\% & \textbf{75\%} & 36\% & 62\% & 83\%\\
DenseNet201 & 77\% & 2\% & 13\% & 66\% & 77\% & \textbf{93\%}\\
GhostNetv3 & 85\% & 17\% & 29\% & 54\% & 58\% & 90\%\\
ShuffleNetv2\_$\times$1.0 & 77\% & 2\% & 13\% & 66\% & 77\% & \textbf{93\%}\\
GhostNet & 66\% & 0\% & 20\% & 45\% & 52\% & 88\%\\
GhostNetv2 & 80\% & 5\% & 28\% & 45\% & 68\% & 87\%\\
\textbf{APSNet} & 90\% & \textbf{33\%} & 71\% & \textbf{88\%} & 81\% & 69\%\\
\botrule
\end{tabular}

\footnotesize{Black bold represents the highest accuracy.}
% \footnotemark{Source: This is an example of table footnote. This is an example of table footnote.}
% \footnotetext[1]{Example for a first table footnote. This is an example of table footnote.}
% \footnotetext[2]{Example for a second table footnote. This is an example of table footnote.}
\end{table*}

\end{appendices}

%%===========================================================================================%%
%% If you are submitting to one of the Nature Portfolio journals, using the eJP submission   %%
%% system, please include the references within the manuscript file itself. You may do this  %%
%% by copying the reference list from your .bbl file, paste it into the main manuscript .tex %%
%% file, and delete the associated \verb+\bibliography+ commands.                            %%
%%===========================================================================================%%

\bibliography{sn-bibliography}% common bib file
%% if required, the content of .bbl file can be included here once bbl is generated
%%\input sn-article.bbl

\end{document}